\documentclass{article}

\usepackage{microtype}
\usepackage{graphicx}
\usepackage{booktabs} 

\usepackage{microtype}
\usepackage{graphicx}
\usepackage{caption}
\usepackage{booktabs} 
\usepackage{tikz}
\usepackage{tablefootnote}
\usepackage{pgfplots, pgfplotstable}
\usepgfplotslibrary{groupplots}
\usepgfplotslibrary{fillbetween}
\usepgfplotslibrary{statistics}
\usepackage{amsmath}
\usepackage{dsfont}
\usepackage{scalefnt}
\usepackage{hyperref}
\usepackage[accepted]{styles/icml2020}

\usepackage{styles/custom}
\usetikzlibrary{positioning,decorations.pathreplacing,decorations.markings,calc,arrows}

\icmltitlerunning{Puzzle Mix: Exploiting Saliency and Local Statistics for Optimal Mixup}

\begin{document}

\twocolumn[
\icmltitle{Puzzle Mix: Exploiting Saliency and Local Statistics for Optimal Mixup}

\begin{icmlauthorlist}
\icmlauthor{Jang-Hyun Kim}{snu,nprc}
\icmlauthor{Wonho Choo}{snu,nprc}
\icmlauthor{Hyun Oh Song}{snu,nprc}
\end{icmlauthorlist}

\icmlaffiliation{snu}{Department of Computer Science and Engineering, Seoul National University, Seoul, Korea}
\icmlaffiliation{nprc}{Neural Processing Research Center}

\icmlcorrespondingauthor{Hyun Oh Song}{hyunoh@snu.ac.kr}

\icmlkeywords{Augmentation, Mixup, Classification, Robustness}

\vskip 0.3in
]

\printAffiliationsAndNotice{}

\begin{abstract}
While deep neural networks achieve great performance on fitting the training distribution, the learned networks are prone to overfitting and are susceptible to adversarial attacks. In this regard, a number of mixup based augmentation methods have been recently proposed. However, these approaches mainly focus on creating previously unseen virtual examples and can sometimes provide misleading supervisory signal to the network. To this end, we propose Puzzle Mix, a mixup method for explicitly utilizing the saliency information and the underlying statistics of the natural examples. This leads to an interesting optimization problem alternating between the multi-label objective for optimal mixing mask and saliency discounted optimal transport objective. Our experiments show Puzzle Mix achieves the state of the art generalization and the adversarial robustness results compared to other mixup methods on CIFAR-100, Tiny-ImageNet, and ImageNet datasets. The source code is available at \url{https://github.com/snu-mllab/PuzzleMix}.
\end{abstract}

\section{Introduction}\label{sec:intro}
Deep neural network models are the bedrock of modern AI tasks such as object recognition, speech, natural language processing, and reinforcement learning. However, these models are known to memorize the training data and make overconfident predictions often resulting in degraded generalization performance on test examples \cite{dropout, zhang2016}. Furthermore, the problem is exacerbated when the models are evaluated on examples under slight distribution shift \cite{ben-david10}. 

To this end, data augmentation approaches aim to alleviate some of these issues by improving the model generalization performance \cite{he15, cutout}. Recently, a line of research called \emph{mixup} has been proposed. These methods mainly focus on creating previously unseen virtual mixup examples via convex combination or local replacement of data for training \cite{mixup, manifoldmixup, cutmix, guo_aaai}. 

\begin{figure}
    \centering
    \includegraphics[width=\columnwidth]{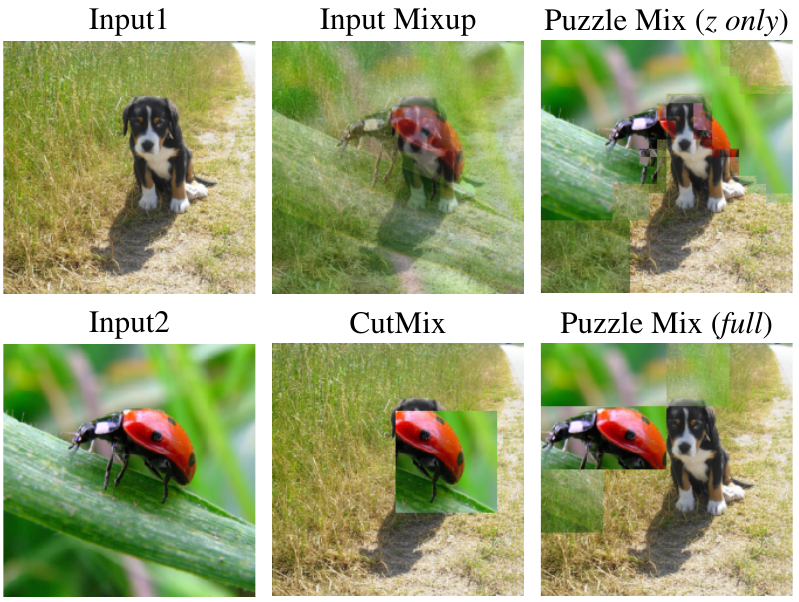}
    \vspace{-0.6cm}
    \caption{A visual comparison of the mixup methods. Puzzle Mix ensures to contain sufficient saliency information while preserving the local statistics of each input.}
    \label{fig:compare}
    \vspace{-0.7em}
\end{figure}

However, the underlying data domains contain rich regional saliency information ($i.e.$ foreground objects in vision, 
prominent syllables in speech, informative textual units in language) \cite{saliency, kalinli2007saliency, Lexrank} and exhibit local regularity structure far from random matrices of numbers \cite{huang1999, speech_recog, discourse}. Thus, completely disregarding these aspects of data could lead to creating mixup examples which could misguide the training model and undermine the generalization performance.

Motivated by this intuition, we propose Puzzle Mix, a mixup method for explicitly leveraging the saliency information and the underlying local statistics of natural examples. Our proposed method jointly seek to find (1) the optimal mask for deciding how much of the two inputs to reveal versus conceal in the given region and for (2) the transport for finding the optimal moves in order to maximize the exposed saliency under the mask. The optimization process is reminiscent of the sliding block puzzle and thus the name Puzzle Mix. Additionally, we impose the objective to respect the various underlying local statistics encouraging the optimization to preserve the structural integrity of each data. The proposed method alternates between finding the optimal mask and optimizing the transport plans, and efficiently generates the mixup examples in a mini-batch stochastic gradient descent setting.

Furthermore, our method allows us to incorporate adversarial training without any computation overhead. Adversarial training is a method for training a robust model resistant to adversarial attacks via optimization \cite{madry}. We adapt the fast adversarial training method from \citet{fast_imagenet} and stochastically include the adversarially perturbed examples with random restarts for robustness.

Our results on CIFAR-100, Tiny-ImageNet, and ImageNet datasets show significant improvement both in the generalization task and in the adversarial robustness over existing mixup methods by a large margin. 

\section{Related Works}\label{sec:rel}

\textbf{Data augmentation}~ Methods that implement data augmentation aim to regularize the models from overfitting to the training distribution and improve the generalization performance by generating virtual training examples in the vicinity of the given training dataset \cite{bishop}. Some of the most commonly used data augmentation techniques are random cropping, horizontal flipping \cite{alexnet}, and adding random noise \cite{noise}.
Recently, a data augmentation method called AugMix is proposed to improve both the generalization performance and the corruption robustness \cite{augmix}. Our method is complementary to these techniques and could be used in conjunction in order to further increase the generalization and robustness performance.

\textbf{Mixup}~ Input mixup creates virtual training examples by linearly interpolating two input data and corresponding one-hot labels \cite{mixup}. The method induces models to have smoother decision boundaries and reduces overfitting to the training data. Manifold mixup extends this concept from input space to feature space \cite{manifoldmixup}. Also, \citet{guo_aaai} proposed an adaptive mixup method, which improves Input mixup by preventing the generation of improper mixup data. \citet{cutmix} proposed CutMix which implants a random rectangular region of the input into another. However, these methods can generate improper examples by randomly removing important regions of the data, which may mislead the neural network (see \Cref{fig:compare}). Our mixup method aims to prevent these issues by utilizing the saliency signal while preserving the local properties of the input data.

\textbf{Saliency}~ 
\citet{saliency} detects object saliency by computing gradients of a pre-trained deep neural network. Subsequently, other methods were introduced to obtain more precise saliency \cite{saliency1, saliency2}. However, these methods require modifying the pre-trained network or training new models to compute the saliency. \citet{cam} and \citet{gradcam} proposed methods with the reduced computation cost but at the cost of saliency resolution. We follow the method from \citet{saliency}, which does not require any modification to the model, to compute the saliency map. The saliency information has been used in various fields of machine learning \citep{saliency_recog, saliency_seg2}.

\textbf{Optimal transport}~
A transport plan that moves a given distribution to another at the minimal cost is called the optimal transport \cite{villani2008optimal}. Also, the optimal transport with discrete domain can be represented as a linear program or an assignment problem \cite{hungarian, villani2008optimal}. The optimal transport problem is widely applied in various applications areas such as color transfer \cite{color} and domain adaptation \cite{otda}. We formulate a binary transport problem for the optimal move, which maximizes the exposed saliency under the mask.

\section{Preliminaries}\label{sec:prelim}
{Let us define $x\in \Xcal$ to be an input data and $y\in \Ycal$ be its output label. Let $\Dcal$ be the distribution over $\Xcal \times \Ycal$. In mixup based data augmentation method, the goal is to optimize the model's loss $\ell: \Xcal \times \Ycal \times \Theta \rightarrow \reals$ given the data mixup function $h(\cdot)$ and the mixing distribution $q$ as below.}\par
\vspace{-1.0em}
\small
\begin{align}\label{eqn:model_opt}
\minimize_\theta \Exp_{(x_0,y_0),(x_1,y_1)\in\Dcal}~ \Exp_{\lambda \sim q} \ell(h(x_0,x_1), g(y_0,y_1); \theta),
\end{align}
\normalsize
{where the label mixup function is $g(y_0,y_1) = (1-\lambda) y_0 + \lambda y_1$. Input mixup uses $h(x_0,x_1) = (1-\lambda) x_0 + \lambda x_1$. Manifold mixup employs $h(x_0,x_1) = (1-\lambda) f(x_0) + \lambda f(x_1)$ for some hidden representation $f$. CutMix defines $h(x_0,x_1) = (1-\mathds{1}_B) \odot x_0 + \mathds{1}_B\odot x_1$ for a binary rectangular mask $\mathds{1}_B$, where $B=[r_x, r_x+r_w]\times [r_y, r_y+r_h]$ with $\lambda=\frac{r_w r_h}{WH}$ and $\odot$ represents the element-wise product. In other words, $B$ is a randomly chosen rectangle covering $\lambda$ proportion of the input. We propose the following mixup function,}\par
\vspace{-1.0em}
\small
\begin{align}
\label{eqn:mixup}
h(x_0,x_1)= (1-z) \odot \Pi_0^\intercal x_0 + z\odot \Pi_1^\intercal x_1,
\end{align}
\normalsize
where $z_i$ represents a mask in $[0,1]$ with mixing ratio $\lambda=\frac{1}{n}\sum_iz_i$. $\Pi_0$ and $\Pi_1$ represent $n\times n$ transportation plans of the corresponding data with $n$ dimensions. $\Pi_{ij}$ encodes how much mass moves from location $i$ to $j$ after the transport. From now on, we omit the dependence of $y$ and $\theta$ from the loss function $\ell$ for clarity. \Cref{tab:mixup_method} summarizes various mixup functions described above. We begin \Cref{sec:method} with the formal desiderata for our mixup function and the corresponding optimization objective.

\begin{table}[t!]
	\centering
	\small
	\begin{tabular}{lc}
	\toprule[1pt]
         	Method & Mixup function $h(x_0, x_1)$\\
                 \midrule
		Input mixup    & $(1-\lambda) x_0+ \lambda x_1$ \\
		Manifold mixup& $(1-\lambda) f(x_0) + \lambda f(x_1)$\\
		CutMix& $(1-\mathds{1}_B) \odot x_0 + \mathds{1}_B\odot x_1$\\
		Puzzle Mix & $\ (1-z) \odot \Pi_0^\intercal x_0 + z\odot \Pi_1^\intercal x_1\ $ \\
		\bottomrule
	\end{tabular}
	\vspace{-0.2em}
	\caption{Summary of various mixup functions.}
	\label{tab:mixup_method}
\end{table}

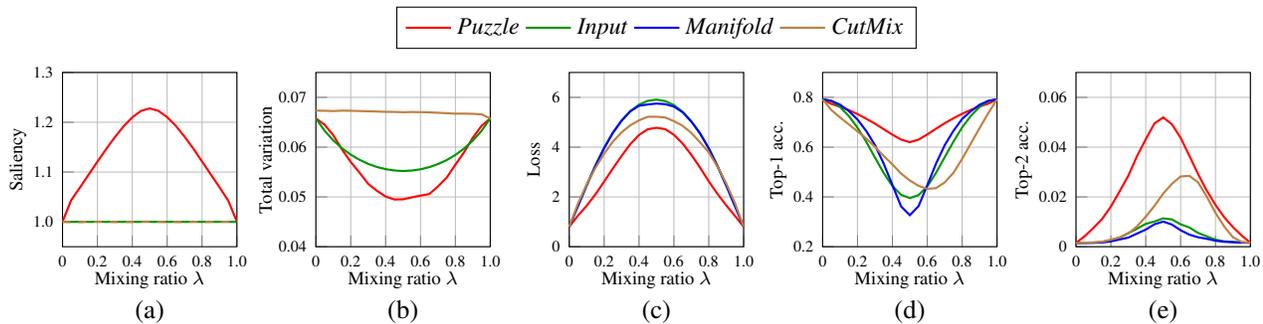
\begin{figure*}[h]
\begin{tikzpicture}[define rgb/.code={\definecolor{mycolor}{RGB}{#1}},
                    rgb color/.style={define rgb={#1},mycolor}]
\begin{groupplot}[
        group style={columns=5, horizontal sep=1.05cm, 
        vertical sep=0.1cm},
		]

\nextgroupplot[
    		width=3.9cm,
    		height=3.9cm,
    		no marks,
    		every axis plot/.append style={thick},
    		grid=major,
    		scaled ticks = false,
    		xlabel near ticks,
    		ylabel near ticks,
    		tick pos=left,
    		tick label style={font=\tiny},
    		xtick={0, 0.2, 0.4, 0.6, 0.8, 1.0},
    		xticklabels={0, 0.2, 0.4, 0.6, 0.8, 1.0},
    		ytick={0.9, 1.0, 1.1, 1.2, 1.3},
    		yticklabels={0.9, 1.0, 1.1, 1.2, 1.3},
    		xlabel shift=-0.18cm,    		
    		ylabel shift=-0.15cm,
    		label style={font=\scriptsize},
    		xlabel={Mixing ratio $\lambda$},
    		ylabel={Saliency},
    		xmin=0,
    		xmax=1,
    		ymin=0.95,
    		ymax=1.3,
            title style={at={(0.5,0)},anchor=north,yshift=-0.8cm},
            title = (a),
            ]
\coordinate (c1) at (rel axis cs:0,1);
\addplot[red] table [x=alpha, y=ours, col sep=comma]{data/saliency.csv};
\addplot[rgb color={0,150,0}] table [x=alpha, y=input, col sep=comma]{data/saliency.csv};
\addplot[brown, dashed] table [x=alpha, y=cutmix, col sep=comma]{data/saliency.csv};

\nextgroupplot[
    		width=3.9cm,
    		height=3.9cm,
    		no marks,
    		every axis plot/.append style={thick},
    		grid=major,
    		scaled ticks = false,
    		xlabel near ticks,
    		ylabel near ticks,
    		tick pos=left,
    		tick label style={font=\tiny},
    		xtick={0, 0.2, 0.4, 0.6, 0.8, 1.0},
    		xticklabels={0, 0.2, 0.4, 0.6, 0.8, 1.0},
    		ytick={0.04, 0.05, 0.06, 0.07, 0.08},
    		yticklabels={0.04, 0.05, 0.06, 0.07, 0.08},
    		xlabel shift=-0.18cm,    		
    		ylabel shift=-0.15cm,
    		label style={font=\scriptsize},
    		xlabel={Mixing ratio $\lambda$},
    		ylabel={Total variation},
    		xmin=0,
    		xmax=1,
    		ymin=0.04,
    		ymax=0.075,
            title style={at={(0.5,0)},anchor=north,yshift=-0.8cm},
            title = (b),
            ]
\addplot[red] table [x=alpha, y=ours, col sep=comma]{data/TV.csv};
\addplot[rgb color={0,150,0}] table [x=alpha, y=input, col sep=comma]{data/TV.csv};
\addplot[brown] table [x=alpha, y=cutmix, col sep=comma]{data/TV.csv};

\nextgroupplot[%
    		width=3.9cm,
    		height=3.9cm,
    		no marks,
    		every axis plot/.append style={thick},
    		grid=major,
    		scaled ticks = false,
    		xlabel near ticks,
    		ylabel near ticks,
    		tick pos=left,
    		tick label style={font=\tiny},
    		xtick={0, 0.2, 0.4, 0.6, 0.8, 1.0},
    		xticklabels={0, 0.2, 0.4, 0.6, 0.8, 1.0},
    		ytick={0, 2, 4, 6, 8},
    		yticklabels={0, 2, 4, 6, 8},
    		xlabel shift=-0.18cm,    		
    		ylabel shift=-0.08cm,
    		label style={font=\scriptsize},
    		xlabel={Mixing ratio $\lambda$},
    		ylabel={Loss},
    		xmin=0,
    		xmax=1,
    		ymin=0,
    		ymax=7,
             title style={at={(0.5,0)},anchor=north,yshift=-0.8cm},
             title = (c),
            ]
\addplot[red] table [x=alpha, y=ours, col sep=comma]{data/loss.csv};
\addplot[rgb color={0,150,0}] table [x=alpha, y=input, col sep=comma]{data/loss.csv};
\addplot[blue] table [x=alpha, y=manifold, col sep=comma]{data/loss.csv};
\addplot[brown] table [x=alpha, y=cutmix, col sep=comma]{data/loss.csv};

\nextgroupplot[%
    		width=3.9cm,
    		height=3.9cm,
    		no marks,
    		every axis plot/.append style={thick},
    		grid=major,
    		scaled ticks = false,
    		xlabel near ticks,
    		ylabel near ticks,
    		tick pos=left,
    		tick label style={font=\tiny},
    		xtick={0, 0.2, 0.4, 0.6, 0.8, 1.0},
    		xticklabels={0, 0.2, 0.4, 0.6, 0.8, 1.0},
    		ytick={0, 0.2, 0.4, 0.6, 0.8, 1.0},
    		yticklabels={0, 0.2, 0.4, 0.6, 0.8, 1.0},
    		xlabel shift=-0.18cm,    		
    		ylabel shift=-0.15cm,
    		label style={font=\scriptsize},
    		xlabel={Mixing ratio $\lambda$},
    		ylabel={Top-1 acc.},
    		xmin=0,
    		xmax=1,
    		ymin=0.2,
    		ymax=0.9,
            legend to name=grouplegend,
    		legend style={legend columns=4, font=\small},
            title style={at={(0.5,0)},anchor=north,yshift=-0.8cm},
            title = (d),
            ]
\addplot[red] table [x=alpha, y=ours, col sep=comma]{data/acc.csv};
\addlegendentry{\textit{Puzzle}}
\addplot[rgb color={0,150,0}] table [x=alpha, y=input, col sep=comma]{data/acc.csv};
\addlegendentry{\textit{Input}}
\addplot[blue] table [x=alpha, y=manifold, col sep=comma]{data/acc.csv};
\addlegendentry{\textit{Manifold}}
\addplot[brown] table [x=alpha, y=cutmix, col sep=comma]{data/acc.csv};
\addlegendentry{\textit{CutMix}}

\nextgroupplot[
    		width=3.9cm,
    		height=3.9cm,
    		no marks,
    		every axis plot/.append style={thick},
    		grid=major,
    		scaled ticks = false,
    		xlabel near ticks,
    		ylabel near ticks,
    		tick pos=left,
    		tick label style={font=\tiny},
    		xtick={0, 0.2, 0.4, 0.6, 0.8, 1.0},
    		xticklabels={0, 0.2, 0.4, 0.6, 0.8, 1.0},
    		ytick={0, 0.02, 0.04, 0.06, 0.08, 0.1},
    		yticklabels={0, 0.02, 0.04, 0.06, 0.08, 0.1},
    		xlabel shift=-0.18cm,    		
    		ylabel shift=-0.15cm,
    		label style={font=\scriptsize},
    		xlabel={Mixing ratio $\lambda$},
    		ylabel={Top-2 acc.},
    		xmin=0,
    		xmax=1,
    		ymin=0,
    		ymax=0.07,
            title style={at={(0.5,0)},anchor=north,yshift=-0.8cm},
            title = (e),
            ]
\coordinate (c2) at (rel axis cs:1,1);
\addplot[red] table [x=alpha, y=ours, col sep=comma]{data/acc2.csv};
\addplot[rgb color={0,150,0}] table [x=alpha, y=input, col sep=comma]{data/acc2.csv};
\addplot[blue] table [x=alpha, y=manifold, col sep=comma]{data/acc2.csv};
\addplot[brown] table [x=alpha, y=cutmix, col sep=comma]{data/acc2.csv};

\end{groupplot}
\coordinate (c3) at ($(c1)!.5!(c2)$);
\node[above] at (c3 |- current bounding box.north) {\pgfplotslegendfromname{grouplegend}};
\end{tikzpicture}
\vspace{-1.0em}
\caption{(a) Mixed saliency $|| h (s(x_0), s(x_1)) ||_1$. Note the saliency map of each input $s(x_k)$ is normalized to sum up to 1. (b) Total variation of mixed data. (c) Cross entropy loss of mixup data and the corresponding soft-label evaluated by the vanilla classifier (ResNet18). (d) Top-1 prediction accuracy of mixed data. Prediction is counted as correct if the Top-1 prediction belongs to $\{y_0, y_1\}$. (e) Top-2 prediction accuracy of mixed data. Prediction is counted as correct if the Top-2 predictions are equal to $\{y_0, y_1\}$. Manifold mixup is omitted in (a) and (b) as manifold mixup generates mixup examples in the hidden space not in the input space.}
\label{fig:loss}
\end{figure*}

\newpage
\section{Methods}\label{sec:method}
Our goal is to maximally utilize the saliency information of each input while respecting the underlying local statistics of the data. First, in order to maximally utilize the saliency information, we seek to find the optimal mixing mask $z$ and the optimal transport plans $\Pi$ under the following criteria.
\vspace{-0.5em}
\begin{itemize}\itemsep=1pt
	\item Given a pair of transported data and a specific region, the mask $z$ should optimally reveal more salient data of the two while masking the less salient one in the given region.
	\item Given a data $x$ and the mask $z$, the transport $\Pi$ should find the optimal moves that would maximize the saliency of the revealed portion of the data. 
\end{itemize}
The criteria above motivates us to maximize for $(1-z) \odot \Pi_0^\intercal s(x_0) + z \odot \Pi_1^\intercal s(x_1)$. Note, we denote the saliency of the input $x$ as $s(x)$ which is computed by taking $\ell_2$ norm of the  gradient values across input channels. \Cref{fig:loss} (a) shows the proposed mixup function well preserves the saliency information after mixup. Second, in order to respect the underlying local statistics of the data \cite{huang1999, speech_recog, discourse}, we consider the following criteria.
\vspace{-0.5em}
\begin{itemize}
	\item The saliency information can be noisy, which could lead to a suboptimal solution. Therefore, we add spatial regularization terms $\psi$ and $\phi_{i,j}$ to control the smoothness of the mask and regional smoothness of the resulting mixed example. \Cref{fig:loss} (b) compares the local smoothness measured in total variation.
	\item We ensure the structural integrity within each data is generally preserved by considering the transport cost $C_{ij}$ (defined as the distance between the locations $i$ and $j$). Also, to further ensure the local salient structure of the data is preserved without being dispersed across after the transport, we optimize for the binary transport plans as opposed to continuous plans.
\end{itemize}
Evaluation results on the pretrained vanilla classifier in \Cref{fig:loss} (c), (d), (e) show our mixup examples have the smallest loss and the highest accuracy compared to other methods, verifying our intuitions above. Moreover, we optimize the main objective after down-sampling the saliency information $s(x)$ with average pooling to support multi-scale transport and masking. From now on, we denote $n$ as the down-sampled dimension. In practice, we select the down-sampling resolution randomly per each mini-batch.

To optimize the mask $z$, we first discretize the range of the mask value. Let $\mathcal{L}$ denote the discretized range $\{\frac{t}{m}\ |\ t=0,1,\dots,m\}$. In addition, to control the mixing ratio of given inputs, we add a prior term $p(z_i)$, which follows a binomial distribution. We now formalize the complete objective in \Cref{eqn:master_eqn}.

\small
\begin{align}
\label{eqn:master_eqn}
    \minimize\limits_{\substack{z\in\mathcal{L}^n\\\Pi_0, \Pi_1 \in \{0,1\}^{n\times n}}}
    & - \norm{(1-z) \odot \Pi_0^\intercal s(x_0) }_1 \\
    & - \norm{z\odot \Pi_1^\intercal s(x_1)}_1 \nonumber\\ 
    &+ \beta\sum_{(i,j) \in \mathcal{N}} \psi(z_i,z_j) + \gamma\sum_{(i,j) \in \mathcal{N}} \phi_{i,j}(z_i,z_j) \nonumber\\
    & - \eta \sum_i \log p(z_i) + \xi\sum_{k=0,1} \langle\Pi_k, C\rangle\nonumber\\
    \mathrm{subject\ to}\ \ &\Pi_k 1_n = 1_n, \ \Pi_k^\intercal 1_n = 1_n \quad \mathrm{for}\ k=0,1.\nonumber
\end{align}
\normalsize

\begin{figure}
    \centering
    \includegraphics[width=0.91\columnwidth]{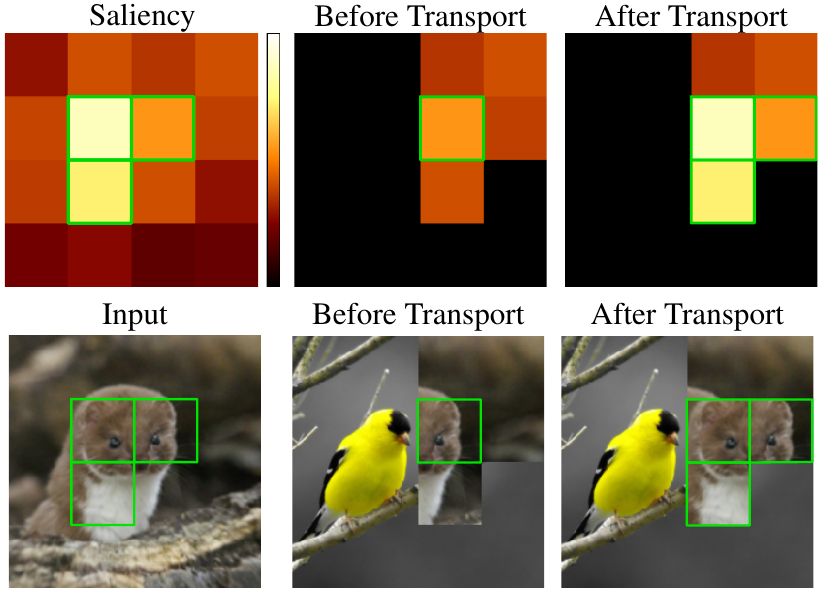}
    \caption{Illustration of Puzzle Mix process. After the transport, the salient regions (highlighted in green) replace the other regions, so that the salient information still remains after the mixup. The first row represents the saliency information after down-sampling, $i.e.$, $s(x)$,  the masked saliency ($z \odot s(x)$), and the masked saliency after transport ($z \odot \Pi^\intercal s(x)$) respectively. The second row shows the corresponding data.}
    \label{fig:transport}
\end{figure}

After solving the optimization problem in \Cref{eqn:master_eqn}, we obtain the mixed example $h(x_0,x_1) = (1-z^*) \odot \Pi_0^{*\intercal} x_0 + z^* \odot \Pi_1^{*\intercal} x_1$ which is then used for the model training as in \Cref{eqn:model_opt}. \Cref{fig:transport} illustrates the mask $z$ and the transport plan $\Pi$ optimized with \Cref{eqn:master_eqn}.

We solve this optimization problem via alternating minimization through iterating first over $z$ and then simultaneously over $\Pi_0$ and $\Pi_1$. In mixup augmentation, however, one needs to be able to efficiently generate the mixed examples as the generation process takes place per each mini-batch. Therefore, we optimize for one complete cycle of the alternating minimization, as repeated cycles require additional network evaluations, for efficiency. As for the initialization, we optimize the mask $z$ with $\Pi_k$ initialized as identity transport, and then optimize each $\Pi_k$ with the previously solved $z$. We now formally discuss individual optimization problems in \Cref{sec:mask_opt} and \Cref{sec:transport_opt}.

\subsection{Optimizing Mask}\label{sec:mask_opt}
Given $\Pi_0$ and $\Pi_1$, we seek to solve the following discrete optimization problem over $z$ in \Cref{eqn:z_opt}. The objective is to decide how to best mix the two transported inputs jointly based on the region saliency measure (unary), the label and data local smoothness (pairwise), and the mixing weight log prior (mix prior) criteria.

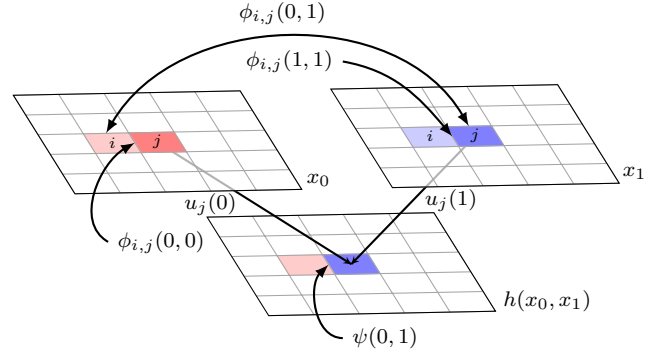
\begin{figure}[t]
\usetikzlibrary{positioning,decorations.pathreplacing,decorations.markings,calc,arrows}
\pgfdeclarelayer{bottom}  \pgfdeclarelayer{middle}
\pgfsetlayers{bottom,middle}  
  \tikzset{myptr/.style={decoration={markings,mark=at position 1 with %
    {\arrow[scale=0.7,>=stealth]{>}}},postaction={decorate}}}
\tikzset{myptr2/.style={decoration={markings,mark=at position 1 with %
    {\arrow[scale=0.5,>=stealth]{>}}},postaction={decorate}}}

   \begin{tikzpicture}[scale=0.8,every node/.style={minimum size=1cm},on grid]
      \begin{pgfonlayer}{bottom}
         \begin{scope}[  
               yshift=0,every node/.append style={
               yslant=0.4,xslant=-1.2,rotate=-35},yslant=0.4,xslant=-1.2,rotate=-35
            ]
            \fill[white,fill opacity=.7] (0.5*5,0.5*0) rectangle (0.5*10,0.5*5);
                        
            \draw[blue!50,fill] (0.5*7,0.5*2) rectangle (0.5*8,0.5*3);
            


            

            \draw[red!20,fill] (0.5*6,0.5*2) rectangle (0.5*7,0.5*3);
            
            
            \draw[step=5mm, gray!70] (0.5*5,0.5*0) grid (0.5*10,0.5*5);
            \draw[black] (0.5*5,0.5*0) rectangle (0.5*10,0.5*5);

            \coordinate (BA) at (0.5*11, 0.5*4);
            \coordinate (BB) at (0.5*12, 0.5*4);
            \coordinate (BC) at (0.5*12, 0.5*3);
            \coordinate (BD) at (0.5*11, 0.5*3);
            \coordinate (BE) at (0.5*7.5, 0.5*2.5);
            \coordinate (B00) at (0.5*10.5, 0.5*3);
            \coordinate (B01) at (0.5*7, 0.5*2.5);
            \coordinate (B11) at (0.5*8.5, 0.5*3);

         \end{scope}
      \end{pgfonlayer}
      \begin{pgfonlayer}{middle}
         \begin{scope}[  
               yshift=50,every node/.append style={
               yslant=0.4,xslant=-1.2,rotate=-35},yslant=0.4,xslant=-1.2,rotate=-35
            ]
            \fill[white,fill opacity=.7] (0.5*1,0.5*1) rectangle (0.5*6,0.5*6);
            
            \draw[red!50,fill] (0.5*3,0.5*3) rectangle (0.5*4,0.5*4);
            



            \draw[red!20,fill] (0.5*2,0.5*3) rectangle (0.5*3,0.5*4);
                                    

            
            \draw[step=5mm, gray!70] (0.5*1,0.5*1) grid (0.5*6,0.5*6);
            \draw[black] (0.5*1,0.5*1) rectangle (0.5*6,0.5*6);
            
            \fill[white,fill opacity=.7] (0.5*8,0.5*1) rectangle (0.5*13,0.5*6);
            
            \draw[blue!50,fill] (0.5*10,0.5*3) rectangle (0.5*11,0.5*4);
            



            \draw[blue!20,fill] (0.5*9,0.5*3) rectangle (0.5*10,0.5*4);
            

            
            \draw[step=5mm, gray!70] (0.5*8,0.5*1) grid (0.5*13,0.5*6);
            \draw[black] (0.5*8,0.5*1) rectangle (0.5*13,0.5*6);
            
            \coordinate (MA) at (0.5*3, 0.5*3.5);
            \coordinate (MB) at (0.5*10, 0.5*3.5);
            \coordinate (MC) at (0.5*14, 0.5*2);
            \coordinate (MD) at (0.5*13, 0.5*2);
            \coordinate (ME) at (0.5*13.5, 0.5*2.5);
            \coordinate (MF) at (0.5*3.5, 0.5*2.5);
            
            \coordinate (ML) at (0.5*2.5, 0.5*3);
            
            \coordinate (MLL) at (0.5*2.5, 0.5*4);
            \coordinate (ML) at (0.5*2.5, 0.5*4);
            \coordinate (MR) at (0.5*10.5, 0.5*4);
            \coordinate (MRL) at (0.5*10.5, 0.5*4);
            
            \coordinate (MLU) at (0.5*3.5, 0.5*3.5);
            \coordinate (MRU) at (0.5*10.5, 0.5*3.5);

         \end{scope}
          \node (a) at (6.8, 2) {\fontsize{8}{9.6}\selectfont {$u_j(1)$}};
          \node (a) at (2.8, 1.93) {\fontsize{8}{9.6}\selectfont{$u_j(0)$}};
          \node (a) at (4, 5.1) {\fontsize{8}{9.6}\selectfont{$\phi_{i, j}(0, 1)$}};
          
          \node (a) at (1.17, 2.95) {\tiny{$i$}};
          \node (a) at (1.92, 2.97) {\tiny{$j$}};
          
          \node (a) at (6.45, 3.06) {\tiny{$i$}};
          \node (a) at (7.2, 3.08) {\tiny{$j$}};          
          
           \node (a) at (4.6, 2.33) {\fontsize{8}{9.6}\selectfont{$x_0$}};
            \node (a) at (9.85, 2.45) {\fontsize{8}{9.6}\selectfont{$x_1$}};
            \node (a) at (8.4, 0.32) {\fontsize{8}{9.6}\selectfont{$h(x_0, x_1)$}};
      \end{pgfonlayer}
      \begin{pgfonlayer}{bottom}
             \draw[myptr, black,thick] (MLU) node[]{}to[] (BE);
             \draw[myptr, black,thick] (MRU) node[]{}to[] (BE);
             
             \draw[-latex,thick] (5., -0.3) node[right]{{\fontsize{8}{9.6}\selectfont{$\psi(0, 1)$}}}to[out=180,in=225] (B01);

      \end{pgfonlayer}
      \begin{pgfonlayer}{middle}
          \draw[-latex,thick] (1.1, 1.3) node[right]{{\fontsize{8}{9.6}\selectfont{$\phi_{i,j}(0, 0)$}}}to[out=150,in=210] (MA);
          \draw[-latex,thick] (5, 4.3) node[left]{{\fontsize{8}{9.6}\selectfont{$\phi_{i,j}(1, 1)$}}}to[out=-10,in=130] (MB);       
          
          \draw[latex-latex,thick] (MR) node[]{}to[out=120,in=60] (ML);

      \end{pgfonlayer}
   \end{tikzpicture}
\vspace{-1.3em}
   \caption{Visualization of different components in the mask optimization. Two rectangles in the top show the two inputs $x_0$ and $x_1$, and the rectangle in the bottom show the mixed output $h(x_0, x_1)$. Figure reproduced with permission from Julien Mairal.}
   \label{fig:graph}
\end{figure}

\small
\begin{align}
\label{eqn:z_opt}
    \minimize\limits_{\substack{z\in\mathcal{L}^n}} 
    & \sum_i u_i(z_i) + \beta\sum_{(i,j) \in \mathcal{N}} \psi(z_i,z_j) \\
    & + \gamma\sum_{(i,j) \in \mathcal{N}} \phi_{i,j}(z_i,z_j) - \eta \sum_i \log p(z_i), \nonumber
\end{align}
\normalsize
where the unary term $u_i(z_i)$ is defined as $z_i(\Pi_0^\intercal s(x_0))_i + (1-z_i)(\Pi_1^\intercal s(x_1))_i$. We define the neighborhood $\mathcal{N}$ as a set of adjacent regions, and use the following pairwise terms and the prior term. \Cref{fig:graph} visualizes different components in \Cref{eqn:z_opt}.

\begin{definition}\label{def:label_smooth}
    (Label smoothness) 
    $\psi(z_i,z_j) \coloneqq (z_i - z_j)^2$.
\end{definition}

For data local smoothness, we measure the distance between input regions. Let $d_p$ denote the distance function. First, we define pairwise terms under the binary case, $\mathcal{L} =\{0,1\}$, and then extend them to the multi-label case. 
\begin{definition}
    (Data local smoothness for binary labels)\\
    Let $x_{k,i}$ represent the $i^{th}$ region of data $x_k$. Then, $\phi^b_{i,j}(z_i,z_j) \coloneqq d_p(x_{z_i, i}, x_{z_j, j})$.
\end{definition}

The discrete optimization problem in \Cref{eqn:z_opt} is a type of multi-label energy minimization problem and can be efficiently solved via $\alpha$-$\beta$ swap algorithm \cite{boykov2001fast}, which is based on the graph-cuts. In the binary label case, finding the minimum s-t cut in the graph returns an equivalent optimal solution if the pairwise term satisfies the submodularity condition \cite{kolmogorov2004energy}. In our problem, the pairwise term is $e_{i,j}(z_i,z_j)=\beta \psi (z_i,z_j) + \gamma \phi_{i,j} (z_i,z_j)$. We now assume that the function values of $d_p$ are bounded in [0,1], which is generally satisfied when data values are bounded in [0,1].

\begin{proposition} Suppose $d_p$ function is bounded in [0,1] and $\phi = \phi^b$. If $\gamma \le \beta$, then $e_{i,j}(z_i,z_j)$ satisfies submodularity for $z_i,z_j \in \{0,1\}$.
\end{proposition}
\begin{proof}
$e(0,0) + e(1,1)$ $= \gamma \phi_{i,j} (0,0) + \gamma\phi_{i,j} (1,1)$ $\le 2\gamma \le 2\beta$
$= \beta \psi (0,1) + \beta \psi (1,0)$ $\le e(0,1) + e(1,0).$
\end{proof}

\begin{figure*}[ht]
\centering
\includegraphics[width=\textwidth]{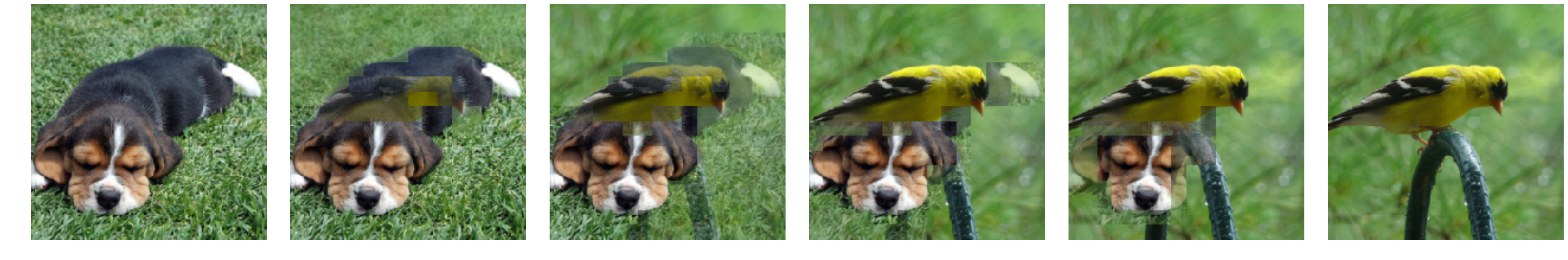}
\includegraphics[width=\textwidth]{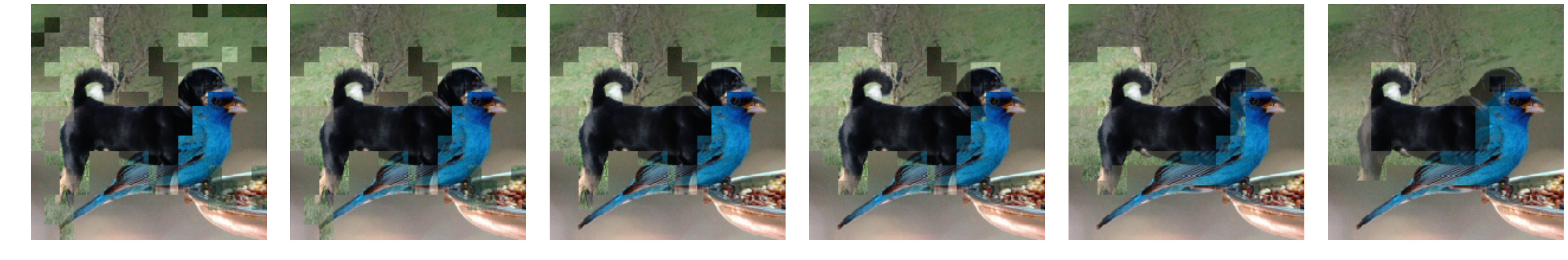}
\caption{(Top row) Puzzle Mix images with increasing mixing weight $\lambda$. (Bottom row) Puzzle Mix images with increasing smoothness coefficients, $\beta$ and $\gamma$. Note that the results are obtained without transport.}
\label{fig:gc_img}
\end{figure*}

For multi-label case, the $\alpha$-$\beta$ swap algorithm iteratively applies graph-cut as a sub-routine and converges to a local-minimum if the pairwise term satisfies pairwise-submodularity \cite{schmidt2011generalized}. We can guarantee pairwise-submodularity by slightly modifying $\phi^b_{i,j}$ as 
\small
\begin{align*}
    &\phi^{b'}_{i,j}(0,0) = \phi^b_{i,j}(0,0) + (\phi^b_{i,j}(0,1) + \phi^b_{i,j}(1,0))/2\\
    &\phi^{b'}_{i,j}(0,1) = \phi^b_{i,j}(0,1) + (\phi^b_{i,j}(0,0) + \phi^b_{i,j}(1,1))/2\\
    &\phi^{b'}_{i,j}(1,0) = \phi^b_{i,j}(1,0) + (\phi^b_{i,j}(0,0) + \phi^b_{i,j}(1,1))/2\\
    &\phi^{b'}_{i,j}(1,1) = \phi^b_{i,j}(1,1) + (\phi^b_{i,j}(0,1) + \phi^b_{i,j}(1,0))/2.
\end{align*}
\normalsize
It is important to note that, $\phi^{b'}_{i,j}(1,0) + \phi^{b'}_{i,j}(0,1) - \phi^{b'}_{i,j}(0,0) - \phi^{b'}_{i,j}(1,1) = 0$.

\begin{definition}\label{def:img_smooth}
    (Data local smoothness for the multi labels)\\
$\phi_{i,j}(z_i,z_j) \coloneqq 
z_i z_j \phi^{b'}_{i,j}(1,1) 
+ z_i (1 - z_j) \phi^{b'}_{i,j}(1,0) 
+ (1-z_i) z_j \phi^{b'}_{i,j}(0,1)
+ (1 - z_i) (1 - z_j) \phi^{b'}_{i,j}(0,0)
,\ ^\forall z_i, z_j \in \mathcal{L}$.
\end{definition}
\vspace{0.4em}

\begin{proposition}
    With $\phi_{i,j}$ defined as  \Cref{def:img_smooth}, $e_{i,j}$ satisfies pairwise submodularity.
\end{proposition}
\vspace{-0.6em}
\begin{proof}
We can represent $\phi_{i,j}$ as follows: 
\small
\begin{align*}
    &\phi_{i,j}(z_i, z_j)\\
    &= f(z_i, z_j) \frac{\phi^{b'}_{i,j}(1,0) + \phi^{b'}_{i,j}(0,1) - \phi^{b'}_{i,j}(0,0) - \phi^{b'}_{i,j}(1,1)}{2} \\
    & + z_i \frac{\phi^{b'}_{i,j}(1,0) + \phi^{b'}_{i,j}(1,1) - \phi^{b'}_{i,j}(0,0) - \phi^{b'}_{i,j}(0,1)}{2} \\
    & + z_j \frac{\phi^{b'}_{i,j}(0,1) + \phi^{b'}_{i,j}(1,1) - \phi^{b'}_{i,j}(0,0) - \phi^{b'}_{i,j}(1,0)}{2} \\
    & + \phi^{b'}_{i,j}(0,0),
\end{align*}
\normalsize
where $f(z_i, z_j) = (1-z_i) z_j + z_i (1 - z_j)$.
By definition $\phi^{b'}_{i,j}(1,0) + \phi^{b'}_{i,j}(0,1) - \phi^{b'}_{i,j}(0,0) - \phi^{b'}_{i,j}(1,1) = 0$, and thus, $\phi_{i,j}(z_i, z_j)$ can be represented as the form of $z_i\phi^{b'_1}_{i,j} + z_j \phi^{b'_2}_{i,j} + c$. Thus, $^\forall x, y \in \mathcal{L},\ \phi_{i,j}(x, y) + \phi_{i,j}(y, x) = x\phi^{b'_1}_{i,j} + y \phi^{b'_2}_{i,j} + c + y\phi^{b'_1}_{i,j} + x \phi^{b'_2}_{i,j} + c = \phi_{i,j}(x, x) + \phi_{i,j}(y, y)$, which means $\phi_{i,j}$ satisfies pairwise submodularity.

By definition $\psi$ satisfies pairwise submodularity, and by \Cref{lem:1}, $e_{i,j}$ satisfies pairwise submodularity. 
\end{proof}

\begin{lemma}\label{lem:1}
    If $\psi$, $\phi$ satisfies pairwise submodularity and $\beta$, $\gamma \in \mathbb{R_+}$, then $\beta \psi + \gamma \phi$ satisfies pairwise submodularity.
    	\vspace{-1em}
\end{lemma}

\begin{proof}
See Supplementary A.1.
    	\vspace{-1em}
\end{proof}

{Finally, we use the prior term to control the ratio of inputs in the mixed output. For the given mixing weight $\lambda$, which represents the ratio of $x_1$ with respect to $x_0$, we define the prior term $p$ to satisfy $\mathbb{E}_{z_i\sim p}[z_i] = \lambda, \forall i$. Specifically, for the label space $\mathcal{L} = \{\frac{t}{m}|\ t=0, ..., m\}$, we define the prior term as $p(z_i=\frac{t}{m}) = \binom{m}{t} \lambda^t (1-\lambda)^{m-t}\ ~\text{for}~\ t=0, 1, ..., m$. In other words, $z_i \sim \frac{1}{m}B(m, \lambda)$.}\par

In \Cref{fig:gc_img}, we provide the resulted mixup images using the optimal mask from \Cref{eqn:z_opt}. Specifically, we visualize how the Puzzle Mix images change by increasing the mixing weight $\lambda$ and the coefficients of the smoothness terms, $\beta$ and $\gamma$.

\subsection{Optimizing Transport}\label{sec:transport_opt}
{After optimizing the mask $z$, we optimize the transportation plans for the input data under the optimal mask $z^*$. Our objective with respect to transportation plans is the following.}\par
\vspace{-1em}
\small
\begin{align*}
    \minimize\limits_{\substack{\Pi_0, \Pi_1 \in \{0,1\}^{n\times n}}} 
    & - \norm{(1-z^*) \odot \Pi_0^\intercal s(x_0) }_1 \\
    & - \norm{(z^*\odot \Pi_1^\intercal s(x_1)}_1 \\
    & + \xi\sum_{k=0,1} \langle\Pi_k, C\rangle\\
    \mathrm{subject\ to}\ \ &\Pi_k 1_n = 1_n, \ \Pi_k^\intercal 1_n = 1_n \quad \mathrm{for}\ k=0,1.
\end{align*}
\normalsize
{Note the problem is completely separable as two independent optimization problems of each $\Pi_k$. Let $s(x_1)_i$ denote the $i^{th}$ entry of the $n$-dimensional column vector $s(x_1)$. The term $\norm{z^* \odot \Pi_1^\intercal s(x_1) }_1$ can be represented as $\sum_{i,j} z^*_j s(x_1)_i \Pi_{1 i,j}$ $= \langle \Pi_1, s(x_1) z^{*\intercal}\rangle$. Finally, the transport optimization problem of $\Pi_1$ becomes}\par
\vspace{-1em}
\small
\begin{align}
\label{eqn:pi_opt}
    \minimize\limits_{\substack{\Pi_1 \in \{0,1\}^{n\times n}}} 
    &\ \langle\Pi_1, C'\rangle\\
    \mathrm{subject\ to}\ \ &\Pi_1 1_n = 1_n, \ \Pi_1^\intercal 1_n = 1_n,\nonumber
\end{align}
\normalsize
where $C' = \xi C - s(x_1) z^{*\intercal}$. $C'_{ij}$ is the cost of moving the $i^{th}$ region to the $j^{th}$ position, which consists of two components. The first component is the distance $\xi C_{ij}$, which is defined as a distance from $i$ to $j$. The second component is the saliency term, which discounts the transport cost with the saliency value of the $i^{th}$ region if the mask of $j^{th}$ position is non-zero.
Briefly speaking, the larger the saliency value, the more the discount in the transport cost. 

The optimization problem in \Cref{eqn:pi_opt} can be solved exactly by using the Hungarian algorithm and its variants with time complexity of $O(n^3)$~\cite{hungarian, jv}. As we need to efficiently generate mixup examples per each mini-batch, this can be a computational bottleneck as $n$ increases. Thus, we propose an approximate algorithm that can be parallelized on GPUs and efficiently computed in batches. The proposed algorithm can quickly decrease the objective $\langle\Pi, C'\rangle$ and converges to a local-minimum within $n(n-1)/2 + 1$ steps. Experimental results comparing the wall clock execution time and the relative error are in Supplementary B.

\begin{algorithm}[t]
    \caption{Masked Transport}
    \label{alg:transport}
\begin{algorithmic}
    \STATE {\bfseries Input:} mask $z^*$, cost $C'$, large value $v$
    \STATE Initialize $C^{(0)} = C',\ t=0$
    \REPEAT
    \STATE $target = \argmin (C^{(t)}, \text{dim}=1)$

    \STATE $\Pi = 0_{n\times n}$
    \FOR{$i=0$ {\bfseries to} $n-1$}
    \STATE $\Pi[i, target[i]] = 1$
    \ENDFOR
    
    \STATE $C_{conflict} = C^{(t)} \odot \Pi + v(1-\Pi) $
    \STATE $source = \argmin (C_{conflict}, \text{dim}=0)$
    
    \STATE $\Pi_{win} = 0_{n\times n}$
    \FOR{$j=0$ {\bfseries to} $n-1$}
    \STATE $\Pi_{win}[source[j], j] = 1$
    \ENDFOR

    \STATE $\Pi_{win} = \Pi_{win} \odot \Pi $
    \STATE $\Pi_{lose} = (1 - \Pi_{win}) \odot \Pi $ 
    \STATE $C^{(t+1)} = C^{(t)} + v\Pi_{lose}$
    \STATE $t = t + 1$
    \UNTIL{convergence}
    \STATE {\bfseries Return:} $\Pi_{win}$
\end{algorithmic}
\end{algorithm}

\Cref{alg:transport} progressively alternates between row-wise and column-wise optimizations. The algorithm first minimizes $\langle\Pi, C'\rangle$ only with the $\Pi 1_n = 1_n$ constraint. However, since the optimization is done without the column constraint, there can be multiple 1 values in a column of $\Pi$. In the following step, the column with multiple 1 values leaves only one 1 in the row with the smallest cost. We denote the result as $\Pi_{win}$ in \Cref{alg:transport}. The corresponding cost entries for the rows that do not remain in $\Pi_{win}$ are penalized with a large additive value, and the 1 values are moved to the other columns in the next iteration.

Our algorithm can also take advantage of intermediate $\Pi_{win}$ as a solution, supported by the following two properties. We suppose that transport cost matrix $C$ has zeros in diagonal entries and positive values in others. In addition, let $\Pi^{(t)}$ and $\Pi^{(t)}_{win}$ denote $\Pi$ and $\Pi_{win}$ at the end of $t^{th}$ step in \Cref{alg:transport}. 

\begin{proposition}\label{prop:win}
     Suppose $z^*$ has values in $\{0,1\}$. Then for $j$ s.t. $z^*_j = 1$, $j^{th}$ column of $\Pi^{(t)}_{win}$ has exactly one 1.
\end{proposition}
\begin{proof}
By the definition of $C^{(0)}=\xi C - s(x_0) z^{*\intercal}$, for $j$ s.t. $z^*_j = 1$, $j^{th}$ row of $C^{(0)}$ has a minimum at $j^{th}$ entry. Thus, $j^{th}$ column of $\Pi^{(0)}_{win}$ has exactly one 1 and others are 0. Suppose that, the claim is satisfied for $\Pi^{(t)}_{win}$ and $\Pi^{(t)}_{win}[i(j),j]$ is 1 for $j$ s.t. $z^*_j = 1$. Then, by the definition of $\Pi^{(t)}_{win}$, $C^{(t)}_{win}[i(j),j]$ is the minimum of $i(j)^{th}$ row of $C^{(t)}$ and the row will not be updated in $C^{(t+1)}$. Thus, $i(j)^{th}$ row of $C^{(t+1)}$ has a minimum at $j^{th}$ entry and $j^{th}$ column of $\Pi^{(t+1)}_{win}$ has exactly one 1. By induction, the claim holds.
\end{proof}

\begin{proposition}
    Under the assumption of \Cref{prop:win}, the partial objective $<\Pi_{win}^{(t)}, C' z^*>$ decreases as t increases.
\end{proposition}
\begin{proof}
By Proposition 3, for $j$ s.t. $z^*_j = 1$, $j^{th}$ column of $\Pi^{(t)}_{win}$ has exactly one 1. Let $i(j; t)$ denote the corresponding row index with the entry 1. Then, it is enough to prove that $C'[i(j; t+1), j] \le C'[i(j; t), j]$. However, in the last part of the proof of Proposition 3, we showed that $i(j;t)^{th}$ row of $C^{(t+1)}$ has a minimum at $j^{th}$ entry, and thus $\Pi^{(t+1)}[i(j; t), j]=1$. By Algorithm 1, index $i(j; t+1)$ satisfies $C^{(t+1)}[i(j; t+1), j] \le C^{(t+1)}[i, j]$, $\forall i$ s.t. $\Pi^{(t+1)}[i, j]=1$. Thus, $C^{(t+1)}[i(j; t+1), j] \le C^{(t+1)}[i(j; t), j]$. Finally, $\Pi^{(t+1)}[i, j]=1$ means that cost from $i$ to $j$ is not updated, $i.e.$, $C^{(t+1)}[i,j] = C'[i,j]$.
\end{proof}

Finally, we introduce the convergence property of \Cref{alg:transport}. 
\begin{proposition}\label{prop:convergence}
    \Cref{alg:transport} converges to a local-minimum with respect to the update rule at most $n(n-1)/2 + 1$ steps.
\end{proposition}
\begin{proof}
See Supplementary A.2.
\end{proof}

\subsection{Adversarial Training}\label{sec:adv}
Since our mix-up strategy utilizes the gradients of the loss function with respect to the given inputs for saliency computation, we can incorporate adversarial training in our mix-up method without \emph{any} additional computation cost.  

\begin{algorithm}[t]
    \caption{Stochastic Adversarial Puzzle Mix}
    \label{alg:adv}
\begin{algorithmic}
    \STATE {\bfseries Input:} data $x_0, x_1$, attack ball $\epsilon$, step $\tau$, probability $p$
    \STATE $x_{i,clean}$ = $x_i\ $ for $i=0,1$
    \STATE Sample $\nu_i \sim B(1, p) \ $ for $i=0,1$
    \FOR{$i = 1,2$}
    \IF{$\nu_i == 1$}
    \STATE $\kappa_i \sim Uniform(-\epsilon, \epsilon)$
    \STATE $x_{i} \leftarrow x_i + \kappa_i$
    \ENDIF
    \ENDFOR

    \STATE Calculate gradient $\nabla_x l(x_i)\ $ for $i = 0,1$
    \STATE Optimize $z^*$ and $\Pi^*_i$ in \Cref{eqn:master_eqn}
    \STATE Sample $\delta \sim Uniform(0, 1)\ $
    \FOR{$i=0,1$}
    \IF{$\nu_i == 1$}
    \STATE $\kappa_i \leftarrow \kappa_i + \tau\ sign(\nabla_x l(x_i))$
    \STATE $\kappa_i \leftarrow clip(\kappa_i, -\epsilon, \epsilon)$
    \STATE $x_i \leftarrow x_{i,clean} + \delta\ \kappa_i$
    \ENDIF
    \ENDFOR

    \STATE {\bfseries Return:} $(1-z^*) \odot \Pi_0^{*\intercal} x_0 + z^* \odot \Pi_1^{*\intercal} x_1$
\end{algorithmic}
\end{algorithm}

For adversarial training on mixup data, we adapt the fast adversarial training method of \citet{fast_imagenet}, which adds a uniform noise before creating an adversarial perturbation. As shown in \Cref{alg:adv}, we add the adversarial perturbation to the proper location of the mixed output, $ i.e.$, adding an adversarial signal to the corresponding input and location specified by $z$. Note that the adversarial perturbation is added to each data probabilistically to prevent possible degradation in the generalization performance. 

\section{Implementation Details}\label{sec:impl}
First, to solve the discrete optimization problem with respect to the mask $z$, we use $\alpha$-$\beta$ swap algorithm from the pyGCO python wrapper\footnote{https://github.com/Borda/pyGCO}. Although the minimization is performed example-wise in CPUs, the $\alpha$-$\beta$ swap algorithm converges quickly, since we restrict the size of the graph with down-sampling. Note that, in our experiments, the computational bottleneck of the method is in the forward-backward passes of the neural network. In our experiments, we use label space $\mathcal {L} = {\{0, \frac{1}{2}, 1\}}$. In addition, we randomly sample the size of the graph, \textit{i.e.}, size of mask $z$, from $\{2\times2,\ 4\times4,\ 8\times8,\ 16\times16\}$, and down-sample the given mini-batch for all experiments. 

We normalize the down-sampled saliency map, which is used as the unary term, to sum up to 1. This allows us to use consistent hyperparameters \emph{across all the} models and datasets.
To measure the distance between the two adjacent data regions, we compute the mean of the absolute values of differences on the boundaries. For the mixing ratio $\lambda$, we randomly sample $\lambda$ from $Beta(\alpha, \alpha)$ at each mini-batch. All of the computations in our algorithm except $\alpha$-$\beta$ swap are done in mini-batch and can be performed in parallel in GPUs. Note that for-loops in Algorithm 1 can be done in parallel by using the scatter function of PyTorch \cite{pytorch}. 

Since we calculate the saliency information by back-propagating the gradient of loss function through the model, we can utilize this gradient information without any computational overhead. We regularize the gradient of the model with mixup data as $\nabla_\theta\ell(h(x_0, x_1), g(y_0, y_1);\theta) + \frac{1}{2}\lambda_{clean}(\nabla_\theta\ell(x_0, y_0;\theta) + \nabla_\theta\ell(x_1, y_1;\theta))$. This additional regularization helps us to improve generalization performance on Tiny-ImageNet and ImageNet.

\section{Experiments}\label{sec:exp}
We train and evaluate classifiers on CIFAR-100 \cite{cifar}, Tiny-ImageNet \cite{tiny-imagenet}, and ImageNet \cite{imagenet} datasets. We first study the generalization performance and adversarial robustness of our method (\Cref{exp:gen}). Next, we show that our method can be used in conjunction with the existing augmentation method (AugMix) to simultaneously improve the corruption robustness and generalization performance (\Cref{exp:robust}). Finally, we perform ablation studies for our method (\Cref{exp:ablation}).

\subsection{Generalization Performance and Adversarial Robustness}\label{exp:gen}
\subsubsection{CIFAR-100}
We train two residual neural networks \cite{he15}: WRN28-10 \cite{wrn} and PreActResNet18 \cite{preactresnet}. We follow the training protocol of \citet{manifoldmixup}, which trains WRN28-10 for 400 epochs and PreActResNet18 for 1200 epochs. Hyperparameter settings are available in Supplementary C.1. We reproduce the mixup baselines \cite{mixup, manifoldmixup, cutmix, augmix} and compare the baselines with our method under the same experimental settings described above. We denote the experiments as \textit{Vanilla, Input, Manifold, CutMix, AugMix, Puzzle Mix} in the experiment tables. 

Note however, our mixup method requires an additional forward-backward evaluation of the network per mini-batch to calculate the saliency signal. For some practitioners, a fairer comparison would be to compare the performances at a fixed number of network evaluations (\ie~ for power conservation). In order to compare our method in this condition, we also test our method trained for half the epochs and with twice the initial learning rate. We denote this experiment as \textit{Puzzle Mix (half)} in the experiment tables.

In addition, we report experiments with the adversarial training described in \Cref{alg:adv} with $p=0.1$. We denote this experiment as \textit{Puzzle Mix (adv)} in the tables. We assess the adversarial robustness against FGSM attack of 8/255 $\ell_\infty$ epsilon ball following the evaluation protocol of \citet{mixup, manifoldmixup, cutmix} for fair comparison. The results are summarized in \Cref{tab:cifar100_wrn} and \Cref{tab:cifar100_pre}, and adversarial robustness results  against the PGD attack \cite{madry} are in Supplementary D.2.

We observe that Puzzle Mix outperforms other mixup baselines in generalization and adversarial robustness with WRN28-10 (\Cref{tab:cifar100_wrn}) and PreActResNet18 (\Cref{tab:cifar100_pre}). With WRN28-10, \textit{Puzzle Mix} improves Top-1 test error over the best performing baseline by 1.45\%, and \textit{Puzzle Mix (half)} outperforms by 1.17\%. \textit{Puzzle Mix (adv)} improves FGSM error rate over 8.41\% than AugMix while achieving 1.39\% lower Top-1 error rate than Manifold mixup, which had the best Top-1 score among baselines. We observe similar results with PreActResNet18. \textit{Puzzle Mix (adv)} reduces the Top-1 error rate by 1.14\% and the FGSM error rate by 12.98\% over baselines.

\begin{table}[h!]
	\centering
	\small
		\begin{tabular}{lccc}
			\toprule[1pt]
            \multirow{2}{*}{\textbf{Method}} & \multicolumn{1}{c}{\multirow{2}{*}{\begin{tabular}[c]{@{}c@{}}\textbf{Top-1}\\ \textbf{{\fontsize{8}{9.6}\selectfont Error(\%)}}\end{tabular}}} & \multicolumn{1}{c}{\multirow{2}{*}{\begin{tabular}[c]{@{}c@{}}\textbf{Top-5}\\ \textbf{{\fontsize{8}{9.6}\selectfont Error(\%)}}\end{tabular}}} & \multicolumn{1}{c}{\multirow{2}{*}{\begin{tabular}[c]{@{}c@{}}\textbf{FGSM}\\ \textbf{{\fontsize{8}{9.6}\selectfont Error(\%)}}\end{tabular}}} \\
            & \multicolumn{1}{c}{} &
            \multicolumn{1}{c}{} & \multicolumn{1}{c}{}\\
            
			\midrule
			Vanilla  & 21.14 &  6.33 & 63.92 \\
			Input    & 18.27 &  4.98 & 56.60 \\
			Manifold & 17.40 &  4.37 & 60.70 \\
			Manifold$\dagger$ & 18.04 & - & -\\			
			CutMix   & 17.50 &  4.69 & 79.34 \\
			AugMix   & 20.44 &  5.74 & 55.59 \\
			\midrule
			Puzzle Mix     & \textbf{15.95} & 3.92   & 63.71 \\
			Puzzle Mix (half)      & 16.23 & \textbf{3.90}  & 66.74 \\
			\midrule
			Puzzle Mix (adv)       & 16.01 & 3.91 & 47.18 \\
			Puzzle Mix (half, adv) & 16.39 & 3.94 & \textbf{46.95} \\
			\bottomrule[1pt]
		\end{tabular}
	\vspace{-0.2em}
	\caption{Top-1 / Top-5 / FGSM error rates on CIFAR-100 dataset of WRN28-10 trained with various mixup methods (400 epochs). $\dagger$ denotes the result reported in the original paper. Top-1 and Top-5 results are median test errors of models in the last 10 epochs.}
	\label{tab:cifar100_wrn}
\end{table}

\begin{table}[h!]
	\centering
	\small
		\begin{tabular}{lccc}
			\toprule[1pt]
            \multirow{2}{*}{\textbf{Method}} & \multicolumn{1}{c}{\multirow{2}{*}{\begin{tabular}[c]{@{}c@{}}\textbf{Top-1}\\ \textbf{{\fontsize{8}{9.6}\selectfont Error(\%)}}\end{tabular}}} & \multicolumn{1}{c}{\multirow{2}{*}{\begin{tabular}[c]{@{}c@{}}\textbf{Top-5}\\ \textbf{{\fontsize{8}{9.6}\selectfont Error(\%)}}\end{tabular}}} & \multicolumn{1}{c}{\multirow{2}{*}{\begin{tabular}[c]{@{}c@{}}\textbf{FGSM}\\ \textbf{{\fontsize{8}{9.6}\selectfont Error(\%)}}\end{tabular}}} \\
            & \multicolumn{1}{c}{} &
            \multicolumn{1}{c}{} & \multicolumn{1}{c}{}\\
            
			\midrule
			Vanilla  & 23.67 & 8.98 & 88.89 \\
			Input    & 23.16 & 7.58 & 70.09 \\
			Manifold & 20.98 & 6.63 & 73.09 \\
			CutMix   & 23.20 & 8.09 & 86.38 \\
			AugMix   & 24.69 & 8.38 & 76.99 \\
			\midrule
			Puzzle Mix     & \textbf{19.62}  & 5.85 & 79.47\\
			Puzzle Mix (half)   & 20.09 & \textbf{5.59} & 75.72\\
			\midrule
			Puzzle Mix (adv)        & 19.84 & 6.11 & \textbf{57.11}\\
			Puzzle Mix (half, adv)  & 19.96 & 6.20 & 59.33\\
			\bottomrule[1pt]
		\end{tabular}
	\vspace{-0.2em}
	\caption{Top-1 / Top-5 / FGSM error rates on CIFAR-100 dataset of PreActResNet18 trained with various mixup methods.}
	\label{tab:cifar100_pre}
\end{table}

\subsubsection {Tiny-ImageNet}
We train PreActResNet18 network on Tiny-ImageNet dataset, which contains 200 classes with 500 training images and 50 test images per class with $64 \times 64$ resolution \cite{tiny-imagenet}. Training settings are described in Supplementary C.2.
\begin{table}[t]
	\centering
	\small
		\begin{tabular}{lccc}
			\toprule[1pt]
            \multirow{2}{*}{\textbf{Method}} & \multicolumn{1}{c}{\multirow{2}{*}{\begin{tabular}[c]{@{}c@{}}\textbf{Top-1}\\ \textbf{{\fontsize{8}{9.6}\selectfont Error(\%)}}\end{tabular}}} & \multicolumn{1}{c}{\multirow{2}{*}{\begin{tabular}[c]{@{}c@{}}\textbf{Top-5}\\ \textbf{{\fontsize{8}{9.6}\selectfont Error(\%)}} \end{tabular}}} & \multicolumn{1}{c}{\multirow{2}{*}{\begin{tabular}[c]{@{}c@{}}\textbf{FGSM}\\ \textbf{{\fontsize{8}{9.6}\selectfont Error(\%)}}\end{tabular}}} \\
            & \multicolumn{1}{c}{} &
            \multicolumn{1}{c}{} & \multicolumn{1}{c}{}\\
            
			\midrule
			Vanilla  & 42.77 & 26.35& 91.85 \\
			Input    & 43.41 & 26.98 & 88.68 \\
			Manifold & 41.99 & 25.88 & 89.25 \\
			Manifold$\dagger$ & 41.30 & 26.41 & -\\
			CutMix   & 43.33 & 24.48 & 87.19 \\
			AugMix   & 44.03 & 25.32 & 90.00 \\
         \midrule
			Puzzle Mix     & \textbf{36.52} & \textbf{18.95} & 92.52 \\
			Puzzle Mix (half)   & 37.64 & 19.37 & 92.57 \\
			\midrule
			Puzzle Mix (adv)    & 38.55 & 20.48 & \textbf{82.07} \\
			Puzzle Mix (half, adv)   & 38.14 & 19.70 &83.91 \\
			\bottomrule[1pt]
		\end{tabular}
	\vspace{-0.2em}
	\caption{Top-1 / Top-5 / FGSM error rates on Tiny-ImageNet dataset for PreActResNet18 trained with various mixup methods.}
	\label{tab:tiny_pre}
\end{table}

As in the CIFAR-100 experiment, Puzzle Mix shows significant performance gains both on the generalization performance and the adversarial robustness compared to other mixup baselines (\Cref{tab:tiny_pre}).

Puzzle Mix trained with the same number of epochs achieves 36.52\% in Top-1 test error, 5.47\% lower than the strongest baseline, and the model trained with same network evaluations (\textit{half}) outperforms the best baseline by 4.35\%.
Puzzle Mix trained with stochastic adversarial method (\textit{adv}) achieves best Top-1 and FGSM error rate ($\epsilon = 4/255$) compared to other mixup baselines providing 3.44\% lower Top-1 error rate and 5.12\% lower FGSM error rate.

\subsubsection{ImageNet}
In ImageNet experiment, we use ResNet-50 to compare the performance. In order to train the model on ImageNet more efficiently, we utilize the cyclic learning rate, and use pre-resized images following the training protocol in \citet{fast_imagenet} which trains models for 100 epochs. Hyperparameter settings are in Supplementary C.3. Consistent with the previous experiments on CIFAR-100 and Tiny-ImageNet, Puzzle Mix shows the best performance in both Top-1 / Top-5 error rate, achieving 0.43\%, 0.24\% improvement each, compared to the best baseline (\Cref{tab:imagenet}). 

\begin{table}[h!]
	\centering
	\small
		\begin{tabular}{lcc}
			\toprule[1pt]
			\multirow{2}{*}{\textbf{Model}} 
			 & \textbf{Top-1} & \textbf{Top-5}  \\
			 & \textbf{{\fontsize{8}{9.6}\selectfont Error(\%)}} & \textbf{{\fontsize{8}{9.6}\selectfont Error(\%)}}  \\

			\midrule
			Vanilla & 24.31 & 7.34 \\
			Input &22.99 & 6.48 \\
			Manifold &23.15 & 6.50  \\
			CutMix & 22.92 & 6.55  \\
			AugMix & 23.25 & 6.70 \\
      \midrule
			Puzzle Mix & \textbf{22.49} & \textbf{6.24} \\
			\bottomrule[1pt]
		\end{tabular}
		
	\vspace{-0.2em}
	\caption{Top-1 / Top-5 error rates on ImageNet on ResNet-50 following the training protocol in \citet{fast_imagenet} (100 epochs).}
	\label{tab:imagenet}
\end{table}

We further test Puzzle Mix according to the experimental setting of CutMix \citep{cutmix} which trains models for 300 epochs and measures the \textbf{best} test accuracy among the training. As shown in \cref{tab:imagenet_cutmix}, Puzzle Mix outperforms baselines consistently.

\begin{table}[h!]
	\centering
	\small
		\begin{tabular}{lcc}
			\toprule[1pt]
			\multirow{2}{*}{\textbf{Model}} 
			 & \textbf{\fontsize{8}{9.6}\selectfont{Best Top-1}} & \textbf{\fontsize{8}{9.6}\selectfont{Best Top-5}}  \\
			 & \textbf{{\fontsize{8}{9.6}\selectfont Error(\%)}} & \textbf{{\fontsize{8}{9.6}\selectfont Error(\%)}}  \\

			\midrule
			Vanilla & 23.68 & 7.05 \\
			Input & 22.58 & 6.40 \\
			Manifold & 22.50 & 6.21  \\
			CutMix & 21.40 & 5.92  \\
      \midrule
			Puzzle Mix & \textbf{21.24} & \textbf{5.71} \\
			\bottomrule[1pt]
		\end{tabular}
		
	\vspace{-0.2em}
	\caption{Best Top-1 / Top-5 error rates on ImageNet on ResNet-50 following the training protocol in \citep{cutmix} (300 epochs).}
	\label{tab:imagenet_cutmix}
\end{table}

\subsection{Robustness Against Corruption}\label{exp:robust}
\citet{augmix} proposed AugMix which performs Input mixup between clean and augmented images to improve robustness against corrupted datasets as well as the generalization performance. AugMix uses Jensen-Shannon divergence (JSD) between network outputs of a clean image and two AugMix images as a consistency loss. However, computing the JSD term requires triple the network evaluations compared to other mixup methods to train the network.

We found that simply using our mixup algorithm between two AugMix images, improves both the generalization and corruption robustness over the training strategy with the JSD objective. Note that our method requires only one additional (versus two) network evaluation per each mini-batch. We denote this experiment setting as \textit{Puzzle Mix (aug)}.

We use CIFAR-100-C dataset \cite{hendrycks2019robustness} to evaluate the corruption robustness. The dataset consists of 19 types of corruption, including noise, blur, weather, and digital corruption types. In \Cref{tab:corruption}, we report average test errors on CIFAR-100-C dataset as well as test errors on the clean CIFAR-100 test dataset. \Cref{tab:corruption} demonstrates that our method using AugMix images improves both the generalization performance and the corruption accuracy by 3.95\% and 2.31\% each over AugMix baseline.

\begin{table}[h!]
	\centering
	\small
		\begin{tabular}{lcc}
			\toprule[1pt]
            \multirow{2}{*}{\textbf{Method}} & \multicolumn{1}{c}{\multirow{2}{*}{\begin{tabular}[c]{@{}c@{}}\textbf{Top-1}\\ \textbf{{\fontsize{8}{9.6}\selectfont Error(\%)}}\end{tabular}}} & \multicolumn{1}{c}{\multirow{2}{*}{\begin{tabular}[c]{@{}c@{}}\textbf{Corruption}\\ \textbf{{\fontsize{8}{9.6}\selectfont Error(\%)}}\end{tabular}}} \\
            & \multicolumn{1}{c}{} &
            \multicolumn{1}{c}{} \\
            
			\midrule
			Vanilla  & 21.14  & 49.08 \\
			AugMix   & 20.45  & 32.22 \\
			\midrule
			Puzzle Mix     & \textbf{15.95} & 42.46 \\
			Puzzle Mix ($aug$)   & 16.50 & \textbf{29.91} \\
            \bottomrule[1pt]
		\end{tabular}
	\vspace{-0.2em}
	\caption{Top-1 / Corruption error rates on CIFAR-100 and CIFAR-100-C on WRN28-10.}
	\label{tab:corruption}
\end{table}

\subsection{Ablation Study}\label{exp:ablation}
The generalization performance of Puzzle Mix stems from saliency-based multi-label masking and transport. We verified the effectiveness of these two factors in comparative experiments on CIFAR-100 with WRN28-10. \Cref{tab:abl_tp} shows that Puzzle Mix with the binary label space (\textit{binary}) has 1.44\% higher Top-1 error rate than multi-label case, and Puzzle Mix without transport (\textit{mask only}) has 0.43\% higher Top-1 error rate than Puzzle Mix with transport.

\begin{table}[h!]
	\centering
	\small
		\begin{tabular}{lcc}
			\toprule[1pt]
            \multirow{2}{*}{\textbf{Method}} & \multicolumn{1}{c}{\multirow{2}{*}{\begin{tabular}[c]{@{}c@{}}\textbf{Top-1}\\ \textbf{{\fontsize{8}{9.6}\selectfont Error(\%)}}\end{tabular}}} & \multicolumn{1}{c}{\multirow{2}{*}{\begin{tabular}[c]{@{}c@{}}\textbf{Top-5}\\ \textbf{{\fontsize{8}{9.6}\selectfont Error(\%)}}\end{tabular}}} \\
            & \multicolumn{1}{c}{} &
            \multicolumn{1}{c}{} \\
            
			\midrule
			Vanilla  & 21.14 & 6.33 \\
			\midrule
			Puzzle Mix  & \textbf{15.95} & 3.92\\
			Puzzle Mix (\textit{binary}) & 17.39  & 4.34 \\
			Puzzle Mix (\textit{mask only})   & 16.38 & \textbf{3.78} \\
            \bottomrule[1pt]
		\end{tabular}
	\vspace{-0.2em}
	\caption{Top-1 / Top-5 rates on CIFAR-100 dataset of WRN28-10 trained with our mixup methods.}
	\label{tab:abl_tp}
\end{table}

We also verify the effects of different factors in stochastic adversarial training. In \Cref{alg:adv}, we add an adversarial perturbation to each data based on each Bernoulli sample $\nu_i$ and apply linear decay with $\delta$ sampled from the uniform distribution. From \Cref{tab:abl_adv}, we observe that using two independent random variables $\nu_0$ and $\nu_1$ (\textit{adv}) has significant improvement in adversarial robustness over using one variable ($\nu_0=\nu_1$). In the absence of linear decaying (\textit{fgsm}), there is improvement in the FGSM error rate of 4.02\%, but the Top-1 error increases by 0.41\%. In all experiments, $p$ is set to 0.1. We use FGSM attack of 8/255 $\ell_\infty$ epsilon-ball and 7-step PGD attack with a 2/255 step size. Additional experiments regarding the effect of $p$ value in adversarial training are available in Supplementary D.1.

\begin{table}[h!]
	\centering
	\small
		\begin{tabular}{lccc}
			\toprule[1pt]
            \multirow{2}{*}{\textbf{Method}} & \multicolumn{1}{c}{\multirow{2}{*}{\begin{tabular}[c]{@{}c@{}}\textbf{Top-1}\\ \textbf{{\fontsize{8}{9.6}\selectfont Error(\%)}}\end{tabular}}} & \multicolumn{1}{c}{\multirow{2}{*}{\begin{tabular}[c]{@{}c@{}}\textbf{FGSM}\\ \textbf{{\fontsize{8}{9.6}\selectfont Error(\%)}}\end{tabular}}} & \multicolumn{1}{c}{\multirow{2}{*}{\begin{tabular}[c]{@{}c@{}}\textbf{PGD}\\ \textbf{{\fontsize{8}{9.6}\selectfont Error(\%)}}\end{tabular}}} \\
            & \multicolumn{1}{c}{} &
            \multicolumn{1}{c}{} & \multicolumn{1}{c}{}\\
            \midrule
			Puzzle Mix (\textit{adv})    & \textbf{16.01} & 47.18 & \textbf{90.18}\\
			Puzzle Mix (\textit{fgsm})          & 16.42	& \textbf{43.16} & 91.19\\
			Puzzle Mix (\textit{$\nu_0$$=$$\nu_1$}) & 16.66	& 65.90  & 94.05\\
			\bottomrule[1pt]
		\end{tabular}
	\vspace{-0.2em}
	\caption{Top-1 / FGSM / PGD error rates on CIFAR-100 dataset of WRN28-10 trained with our adversarial mixup methods.}
	\label{tab:abl_adv}
\end{table}

\section{Conclusion}\label{sec:conclusion}
We have presented Puzzle Mix, a mixup augmentation method for optimally leveraging the saliency information while respecting the underlying local statistics of the data. Puzzle Mix efficiently generates the mixup examples in a mini-batch stochastic gradient descent setting and outperforms other mixup baseline methods both in the generalization performance and the robustness against adversarial perturbations and data corruption by a large margin on CIFAR-100, Tiny-ImageNet, and ImageNet datasets.

\section*{Acknowledgements}
This research was supported in part by Samsung Electronics and Institute of Information \& communications Technology Planning \& Evaluation (IITP) grant funded by the Korea government (MSIT) (No. 2020-0-00882, (SW STAR LAB) Development of deployable learning intelligence via self-sustainable and trustworthy machine learning). Hyun Oh Song is the corresponding author.

\appendix
\section{Proofs}
\setcounter{lemma}{0}
\subsection{Proof of Lemma1}
\begin{lemma}
  If $\psi$, $\phi$ satisfy pairwise submodularity and $\beta$, $\gamma \in \mathbb{R_+}$, then $\beta \psi + \gamma \phi$ satisfies pairwise submodularity.
\end{lemma}
\begin{proof} By the assumption, $\beta$, $\gamma \in \mathbb{R_+}$, and by using pairwise submodularity of $\psi$ and $\phi$, $\forall x,y \in \mathcal{L}$,
\begin{align*}
&(\beta \psi + \gamma \phi)(x,x) + (\beta \psi + \gamma \phi)(y,y) \\
&= \beta (\psi (x,x) + \psi (y,y)) + \gamma (\phi (x,x) + \phi (y,y))\\
&\le \beta (\psi (x,y) + \psi (y,x)) + \gamma (\phi (x,y) + \phi (y,x))\\
&= (\beta \psi + \gamma \phi)(x,y) + (\beta \psi + \gamma \phi)(y,x).
\end{align*}
\end{proof}

\setcounter{proposition}{4}
\subsection{Proof of Proposition 5}
\begin{proposition}
  Algorithm 1 converges to a local-minimum with respect to the update rule at most $n(n-1)/2 + 1$ steps.
\end{proposition}
\begin{proof}
Let $C^{(0)}$ has a minimum at $(i_1, j_1)$. Then, $\forall t=0,1,...$, $\Pi^{(t)}_{win}[i_1,j_1] = 1$. Next, let's define $I_2 = \{(i,j)|i\ne i_1, j\ne j_1 \}$ and $(i_2,j_2) = \argmin_{(i,j)\in I_2}$ $C^{(1)}[i,j]$. If $\Pi^{(0)}[i_2,j_1] = 1$, $C^{(0)}[i_2,j_1]$ will be added by the large value, and thus, $\Pi^{(1)}_{win}[i_2,j_2] = 1$. Otherwise, if $\Pi^{(0)}[i_2,j_1] = 0$, then $C^{(0)}[i_2,j_2] \le C^{(0)}[i_2,j_1]$ and $\Pi^{(0)}_{win}[i_2,j_2] = 1$. By the definition of $(i_2,j_2)$, $\forall t\ge 1$, $\Pi^{(t)}_{win}[i_2,j_2] = 1$. \\
To use induction, let's define $I_k = \{(i,j)|i\notin \{i_1,..., i_{k-1}\}, j\notin \{j_1,..., j_{k-1}\} \}$ and let $a_{k-1}$ denote a minimal step number at which $\Pi^{(t)}_{win}[i_{k-1}, j_{k-1}]=1$, $\forall t\ge a_{k-1}$. Let's define $(i_k,j_k) = \argmin_{(i,j)\in I_k}$ $C^{(a_{k-1}+1)}[i,j]$. If $\exists j \in \{j_1,..., j_{k-1}\}$, $\Pi_{win}^{(a_{k-1})}[i_k,j] = 1$, then $\forall t\ge a_{k-1}+k-1$, $\Pi^{(t)}_{win}[i_k, j_k]=1$. If not, $\Pi_{win}^{(a_{k-1})}[i_k,j_k] = 1$, and $\forall t\ge a_{k-1}$, $\Pi_{win}^{(t)}[i_k,j_k] = 1$. Thus, $a_k \le a_{k-1} + k-1$.\\
Finally, by induction, $a_{n} \le n(n-1)/2$ which means there are no more updates of $\Pi_{win}$ after $n(n-1)/2 +1$ steps. 
\end{proof}

\section{Analysis of Algorithms}
\subsection{Comparison Experiments for Algorithm 1}\label{sec:sim}
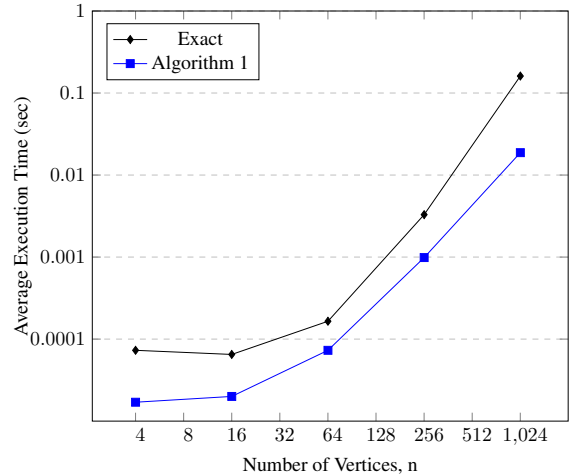
\begin{figure}[h] 
  \scalebox{1.0}{

\pgfplotsset{
  log y ticks with fixed point/.style={
      yticklabel={
        \pgfkeys{/pgf/fpu=true}
        \pgfmathparse{exp(\tick)}%
        \pgfmathprintnumber[fixed relative, precision=3]{\pgfmathresult}
        \pgfkeys{/pgf/fpu=false}
      }
  }
}

\begin{tikzpicture}[scale=0.75]
    \begin{loglogaxis}[
    log ticks with fixed point,
    log basis x=2,
    log basis y=10,
    xlabel={Number of Vertices, n},
    ylabel={Average Execution Time (sec)},
    xticklabel={
        \pgfkeys{/pgf/fpu=true}
        \pgfmathparse{int(2^\tick)}
        \pgfmathprintnumber[fixed]{\pgfmathresult}
    },
    ytick={1e-4, 1e-3, 1e-2, 1e-1, 1e0},
    legend pos=north west,
    ymajorgrids=true,
    grid style=dashed,
    scale only axis,
    xmin=0, xmax=2040,
    ymin=0.00001, ymax=1,
]
 \addplot[
    color=black,
    mark=diamond*,
    ]
    coordinates {
    (4, 0.000073)(16, 0.000065)(64, 0.000165)(256, 0.003291)(1024, 0.161080)
    };

\addplot[
    color=blue,
    mark=square*,
    ]
    coordinates {
    (4, 0.000017)(16, 0.000020)(64, 0.000073)(256, 0.000986)(1024, 0.018754)
    };

\legend{Exact, Algorithm 1}
    \end{loglogaxis}

\end{tikzpicture}

}
  \caption{Comparison of average execution time (log-log scale) to solve Equation (5) between the exact solver (black) and Algorithm 1 (blue). Execution times are mean of 100 trials.}
  \label{fig:time}
\end{figure}

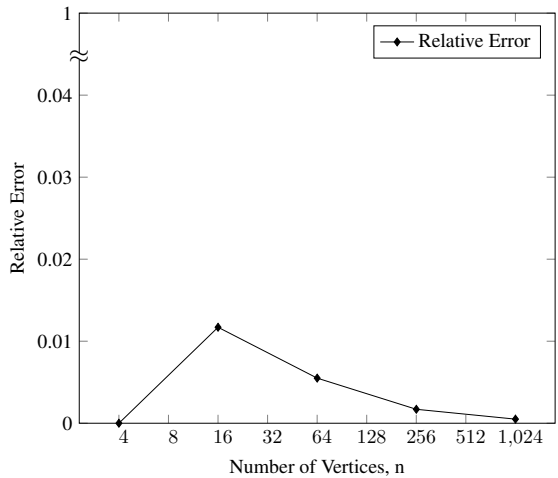
\begin{figure}[h!]
  \hspace{0.4em}\scalebox{1.0}{\begin{tikzpicture}[scale=0.75]
    \begin{semilogxaxis}[
    log ticks with fixed point,
    log basis x=2,
    xlabel={Number of Vertices, n},
    ylabel=Relative Error,
    xticklabel={
        \pgfkeys{/pgf/fpu=true}
        \pgfmathparse{int(2^\tick)}
        \pgfmathprintnumber[fixed]{\pgfmathresult}
    },
    ytick={0, 0.2, 0.4, 0.6, 0.8, 1},
    yticklabels={0, 0.01, 0.02, 0.03, 0.04, 1},
    legend pos=north east,
    scale only axis,
    ylabel near ticks,
    log ticks with fixed point,
    log basis x=2,
    ymin=0, ymax=1,
]

 \addplot[
    color=black,
    mark=diamond*,
    ]
    coordinates {
    (4, 0)(16, 0.234)(64, 0.11)(256, 0.034)(1024, 0.01)
    };
\legend{Relative Error}

\end{semilogxaxis}        
\draw (0, 6.5) -- node[fill=white,inner sep=-1.25pt,outer sep=0,anchor=center]{$\approx$} (0, 6.5);

\end{tikzpicture}}
  \caption{Relative errors of objective function value $f$ between Algorithm 1 (\textit{alg}) and random assignment (\textit{random}). Relative error is calculated as $e_a / (e_a+e_r)$, where $e_a=f_{alg}-f_{exact}, e_r=f_{random}-f_{exact}$.}
  \label{fig:error}
\end{figure}

\Cref{fig:time} and \Cref{fig:error} show the comparison results of Algorithm 1 with the exact Hungarian algorithm on 100 random samples per each vertex size $n$. Note that, the size of a transport plan is $n\times n$. In the simulation, we generate a random cost matrix $C' = C - s(x)z^{\intercal}$, where $s(x)$ is sampled from a uniform distribution and the mask $z$ is sampled from a bernoulli distribution with probability $p=0.5$. In the case of $n=1024$, Algorithm 1 is about 8.6 times faster than the exact algorithm, with relative error of 0.0005. For comparison, we use lapjv solver from library\footnote{https://github.com/berhane/LAP-solvers} as the exact optimizer, which to the best of our knowledge is the fastest solver.

In addition to the simulation test, we train classifiers with ten different random seeds to compare the Top-1 accuracy between using the Hungarian algorithm and Algorithm 1. In this experiment, we train WRN28-10 on CIFAR100 for 400 epochs. In summary, the mean difference of the Top-1 accuracy is $-$0.025, with a standard deviation of 0.239. Besides, we perform a two-sided paired t-test to check the statistical insignificance. As a result, T-statistics is -0.105 with a P-value of 0.919, which means there is no evidence for a statistical difference between the two methods.

\subsection{Convergence of Alternating Minimization}
We solve the optimization problem of a mask and transport plans via alternating minimization for one-cycle for computational efficiency. 
In this subsection, we analyze the convergence property of the alternating algorithm by optimizing 1,000 CIFAR100 image pairs with various numbers of regions. As a result, we observe that the most of optimal masks of the images are not changed after the first cycle (\textit{i.e.,} comparison between one-cycle and multi-cycle), and hence the optimal transport plans are not changed.  For transport cost coefficient $\xi$ in [0.5, 0.8], which shows the best performance, less than 0.2\% of the final mixed images change after the first cycle. Also, the ratio of pixels changed after the first cycle is less than 0.001\%. 

We believe that this result is due to the mutual complement of the optimal mask and the optimal transport. That is, the optimal mask assigns the output regions so that the remained saliency of each input is maximized, and the transport enhances the assignment. Therefore, after a cycle, there is little room for a change of the optimal mask when the optimization is performed again.

\section{Hyperparameter Settings}\label{sec:hyperparam}
\subsection{CIFAR-100}
We train models via stochastic gradient descent (SGD) with initial learning of 0.1 decayed by factor 0.1 at epochs 200 and 300 for WRN28-10 and epochs 400 and 800 for PreActResNet18. We set the momentum as 0.9 and add a weight decay of 0.0001. Mixing weight $\lambda$ is randomly sampled from $Beta(1,1)$ for all experiments except Manifold mixup, which uses $Beta(2,2)$ in the original paper. Puzzle Mix has hyperparameters of $\beta$ for the label smoothness term, $\gamma$ for the data smoothness term, $\eta$ for the prior term, and $\xi$ for the transport cost. In the CIFAR-100 experiment, we use $(\beta, \gamma, \eta, \xi) = (1.2, 0.5, 0.2, 0.8) $. For adversarial training, we use 10/255 epsilon-ball with the step size $\tau$ of 12/255 according to the step size protocol of \citet{fast_imagenet}.

\subsection{Tiny-ImageNet}
We follow the training protocol of \citet{manifoldmixup} except for the learning schedule. \citet{manifoldmixup} train Tiny-ImageNet for 2000 epochs with an initial learning rate of 0.1, but we train models for 1200 epochs with an initial learning rate of 0.2. As in the CIFAR-100 experiment, we use SGD and decay learning rate by factor 0.1 at epochs 600 and 900. We set momentum as 0.9 and weight decay as 0.0001. In case of mixing weight $\lambda$, for Input mixup and Manifold mixup, we follow the setting $\alpha=0.2$ as described in Manifold mixup \cite{manifoldmixup}. For CutMix, we choose $\alpha=0.2$, which showed the best performance among [0.2, 0.5, 1.0], and for Puzzle Mix, we use $\alpha=1.0$. In the Tiny-ImageNet experiment, we use $(\beta, \gamma, \eta, \xi) = (1.2, 0.5, 0.2, 0.8)$, which is the same with the CIFAR-100 experiment. However, we apply regularization using clean input with $\lambda_{clean}=1$ for all experiments regarding Puzzle Mix, and use the same initial learning rate of 0.2 for \textit{Puzzle Mix (half)}.

\subsection{ImageNet}
For ImageNet, we modify the training protocol in \citet{fast_imagenet} and train models for 100 epochs. The learning rate starts from 0.5, linearly increases to 1.0 for 8 epochs, and linearly decreases to 0.125 until 15$^{\text{th}}$ epoch.
Then, the learning rate jumps to 0.2 and linearly decreases to 0.02 until 40$^{\text{th}}$ epoch, 0.002 until 65$^{\text{th}}$ epoch, 0.0002 until 90$^{\text{th}}$ epoch, and 0.00002 until 100$^{\text{th}}$ epoch. In addition, we resize images to 160$\times$160 for the first 15 epochs, and use images pre-resized to 352$\times$352 for the next 85 epochs following the prescription in \citet{fast_imagenet}.
Mixing distribution parameter $\alpha$ is 0.2, 0.2, 1.0, 1.0 each for Input mixup, Manifold mixup, CutMix, Puzzle Mix, which follows the settings of the original papers. In the case of Manifold mixup, there is no experiments on ImageNet, and thus, we tune $\alpha$ in [0.2, 1.0] and report the best result. In the case of ImageNet, we use hyperparameter $(\beta, \gamma, \eta, \xi) = (1.5, 0.5, 0.2, 0.8)$ and apply clean input regularization of $\lambda_{clean} = 1$ for the first 40 epochs. For the experiment following the experimental setting of CutMix \citep{cutmix}, we use the same hyperparameter without the clean input regularization.

\subsection{Hyperparameter Sensitivity}
We analyze the sensitivity of the hyperparameters with WRN28-10 on CIFAR100 trained for 400 epochs. In detail, we sweep hyperparameters one by one while others being fixed, and calculate the mean and standard deviation of Top-1 accuracy. Note that, the hyperparameter setting ($\beta, \gamma, \eta, \xi$) of the main experiment is (1.2, 0.5, 0.2, 0.8) which achieves 15.95\% Top-1 test error.

\Cref{tab:sweep} shows the mean Top-1 error rates and standard deviations of various hyperparameter settings. From the table, we can find that there exists a well of hyperparameters of which performance is superior to that of baselines (manifold mixup: 17.40\%).

\begin{table}[h!]
\centering
\small
\begin{tabular}{lcc}
\toprule[1pt]
\multicolumn{1}{c}{\textbf{Parameter}} & \multicolumn{1}{c}{\textbf{Range}} & \multicolumn{1}{c}{\textbf{Mean Top-1 Error(\%) (SD)}} \\
\midrule
$\beta$      & {[}0.8, 1.6{]}& 16.19\% (0.22)      \\
$\gamma$     & {[}0.0, 1.0{]}& 16.43\% (0.20)       \\
$\eta$       & {[}0.1, 0.35{]} & 16.37\% (0.18)      \\
$\xi$        & {[}0.4, 1.0{]} & 16.25\% (0.27)      \\
\bottomrule[1pt]
\end{tabular}
\caption{Mean Top-1 error rates and standard deviations (SD) for various hyperparameter settings on CIFAR 100 with WRN28-10. For $\beta$, $\gamma$, and $\xi$, we sweep the range with 0.1 step size, and for $\eta$, we sweep the range with 0.05 step size.}
\label{tab:sweep}
\end{table}

\section{Effect of Adversarial Training}\label{sec:adv_effect}
\subsection{Trade-off between Generalization and Adversarial Robustness}
Since adversarial training increases the adversarial robustness at the expense of clean accuracy \cite{madry}, we introduced adversarial probability $p$, a probability of whether to add adversarial perturbation or not, to control the intensity of adversarial training. \Cref{tab:adv_p} shows the inverse relationship between clean error and FGSM error.

\begin{table}[h!]
	\centering
	\small
		\begin{tabular}{cccc}
			\toprule[1pt]
       \multicolumn{1}{c}{\multirow{2}{*}{\begin{tabular}[c]{@{}c@{}}\textbf{Adversarial}\\ \textbf{Probability}\end{tabular}}}& \multicolumn{1}{c}{\multirow{2}{*}{\begin{tabular}[c]{@{}c@{}}\textbf{Top-1}\\ \textbf{Error(\%)}\end{tabular}}} & \multicolumn{1}{c}{\multirow{2}{*}{\begin{tabular}[c]{@{}c@{}}\textbf{Top-5}\\ \textbf{Error(\%)}\end{tabular}}} & \multicolumn{1}{c}{\multirow{2}{*}{\begin{tabular}[c]{@{}c@{}}\textbf{FGSM}\\ \textbf{Error(\%)}\end{tabular}}} \\
      & \multicolumn{1}{c}{} &
      \multicolumn{1}{c}{} & \multicolumn{1}{c}{}\\
       
			\midrule
			0.00 & 37.58 & 19.40 & 92.70 \\
      		0.05 & 37.60 & 19.10 & 89.12 \\
 			0.10 & 37.16 & 19.25 & 86.09 \\
 			0.15 & 38.14 & 19.70 & 83.91 \\
			0.20 & 38.65 & 20.01 & 82.25 \\
			0.25 & 39.46 & 20.40 & 80.37 \\
			0.30 & 40.52 & 21.47 & 79.76 \\
			\bottomrule[1pt]
		\end{tabular}
	\vspace{-0.2em}
	\caption{Top-1 / Top-5 / FGSM error rates on Tiny-ImageNet dataset for PreActResNet18 trained with various adversarial probability $p$.}
	\label{tab:adv_p}
\end{table}

\subsection{Robustness against PGD Attack}
\begin{figure}[h] 
\begin{tikzpicture}[define rgb/.code={\definecolor{mycolor}{RGB}{#1}},
          rgb color/.style={define rgb={#1},mycolor}]
\begin{axis}[
		width=8.2cm,
		height=6.5cm,
		no marks,
		every axis plot/.append style={thick},
		grid=major,
		scaled ticks = false,
		ylabel near ticks,
		tick pos=left,
		tick label style={font=\small},
		xtick={0, 5, 10, 15, 20},
		xticklabels={0, 5, 10, 15, 20},
		ytick={0, 10, 20, 30, 40, 50},
		yticklabels={0, 10, 20, 30, 40, 50},
		label style={font=\small},
		xlabel={PGD iteration},
		ylabel={Top-1 Accuracy},
		xmin=0,
		xmax=20,
		ymin=0,
		ymax=50,
    legend style={nodes={scale=0.7, transform shape}},
		]

\addplot[red] table [x=iter, y=ours, col sep=comma]{data/pgd.csv};
\addlegendentry{$Puzzle$}
\addplot[black] table [x=iter, y=vanilla, col sep=comma]{data/pgd.csv};
\addlegendentry{$Vanilla$}
\addplot[blue] table [x=iter, y=input, col sep=comma]{data/pgd.csv};
\addlegendentry{$Input$}
\addplot[orange] table [x=iter, y=manifold, col sep=comma]{data/pgd.csv};
\addlegendentry{$Manifold$}
\addplot[yellow] table [x=iter, y=cutmix, col sep=comma]{data/pgd.csv};
\addlegendentry{$CutMix$}
\addplot[rgb color={0,100,0}] table [x=iter, y=augmix, col sep=comma]{data/pgd.csv};
\addlegendentry{$AugMix$}
\end{axis}
\end{tikzpicture}
\caption{Adversarial robustness of various mixup methods against the PGD attack.}
\label{fig:pgd}
\end{figure}
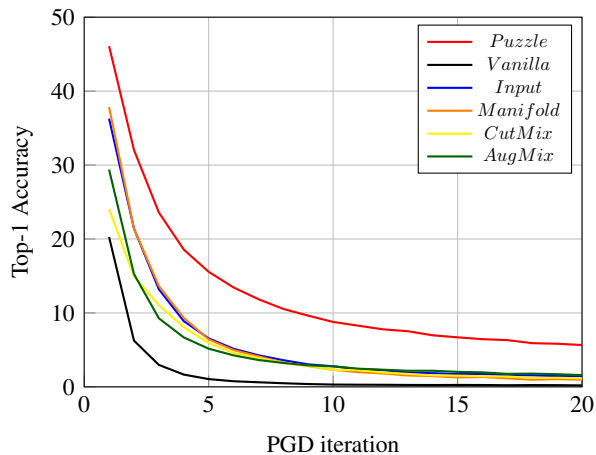

We test the adversarial robustness of various mixup methods against the PGD attack \cite{madry}. In this experiment, we train PreActResNet18 on CIFAR-100 with each mixup method and test with PGD attack of 4/255 $l_\infty$ epsilon-ball with 2/255 step size. For comparison, we test Puzzle Mix with stochastic adversarial training of $p=0.1$, which outperforms other baselines at Top-1 Accuracy given a clean test dataset. \Cref{fig:pgd} demonstrates that Puzzle Mix is more robust against the PGD attack than the existing mixup methods.

\section{Puzzle Mix Qualitative Results}
\subsection{Effect of Prior and Smoothness Term}
\begin{figure*}[t]
\centering
\includegraphics[width=\textwidth]{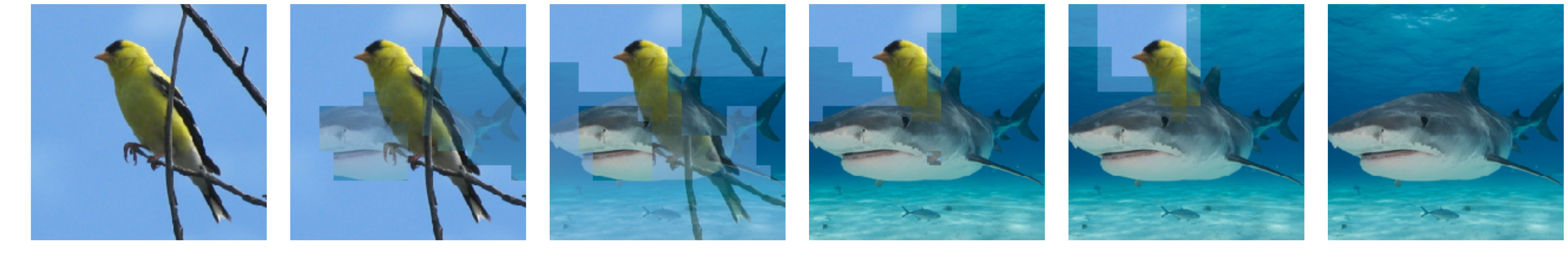}
\includegraphics[width=\textwidth]{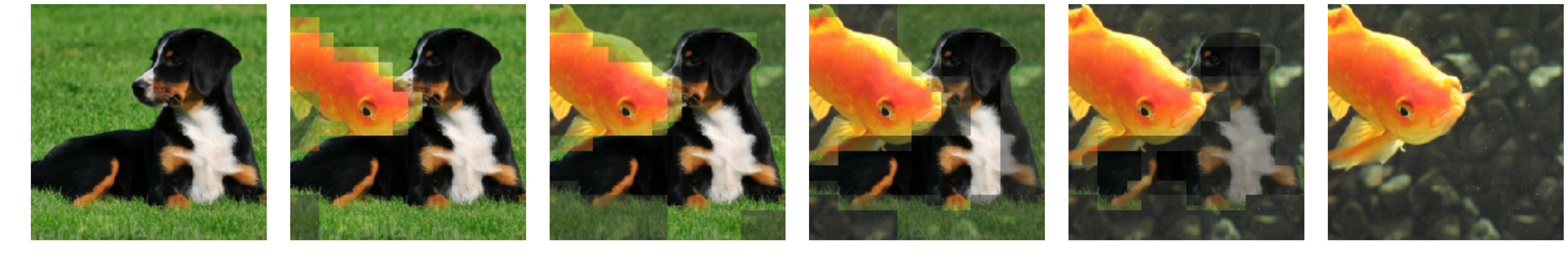}
\includegraphics[width=\textwidth]{figures/alpha/alpha3.pdf}
\includegraphics[width=\textwidth]{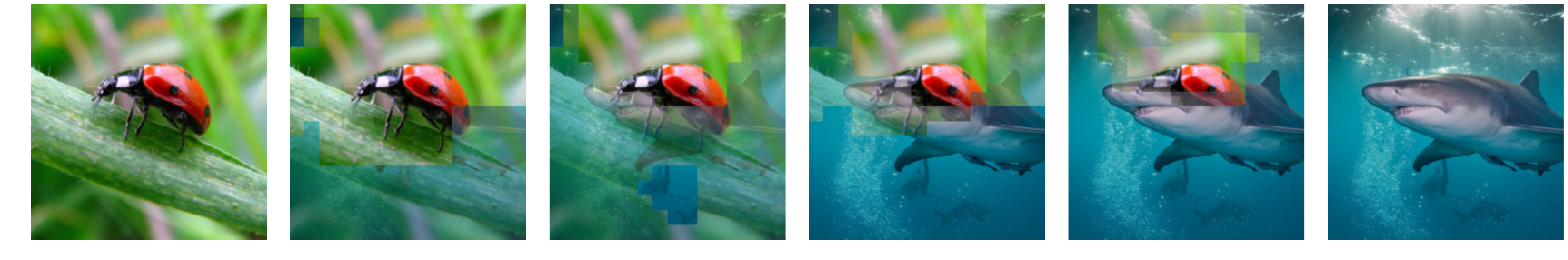}
\includegraphics[width=\textwidth]{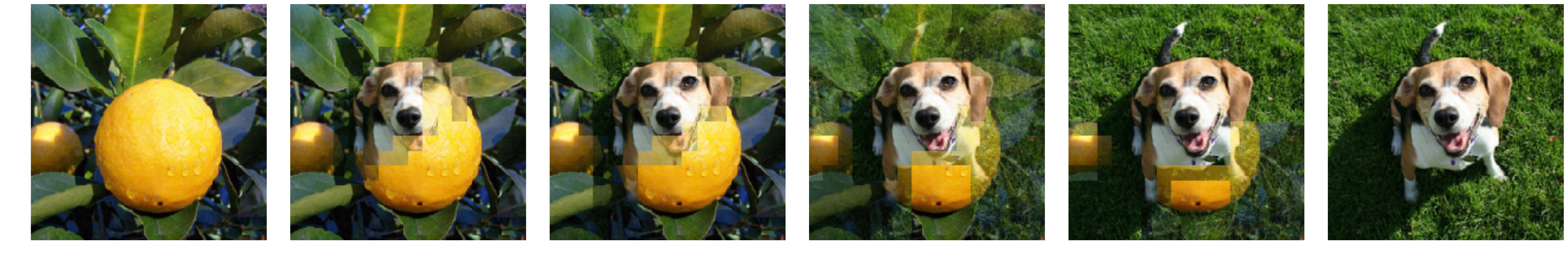}
\includegraphics[width=\textwidth]{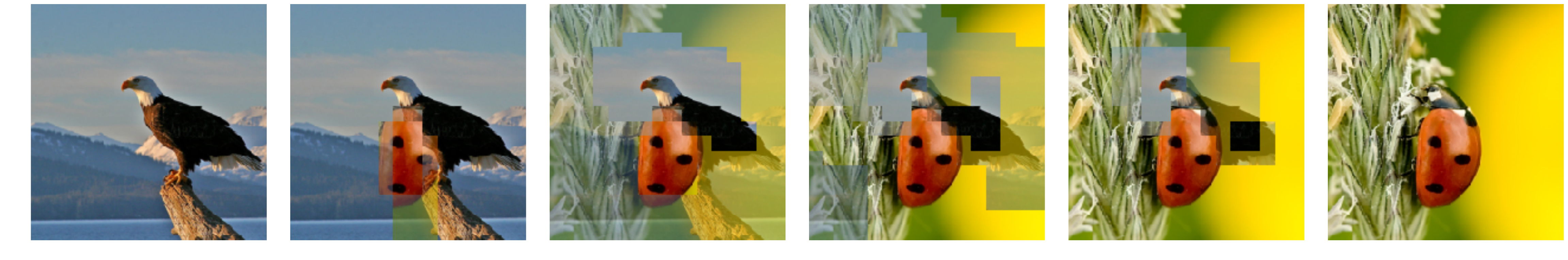}
\includegraphics[width=\textwidth]{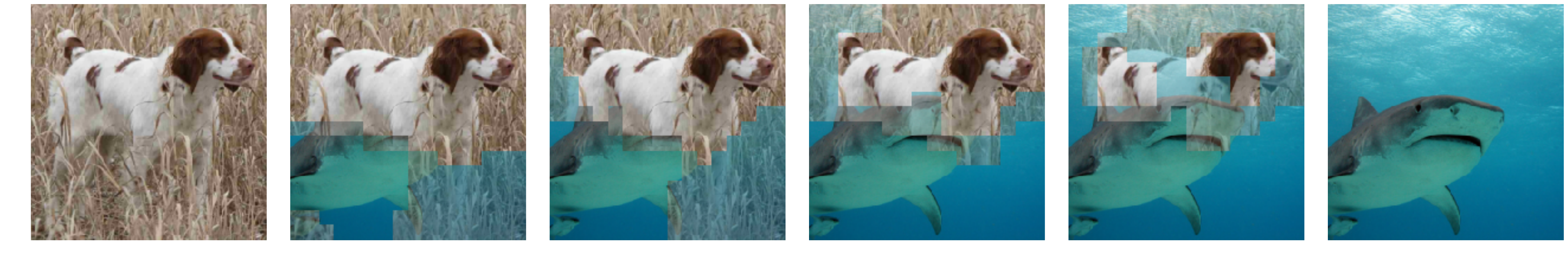}
\includegraphics[width=\textwidth]{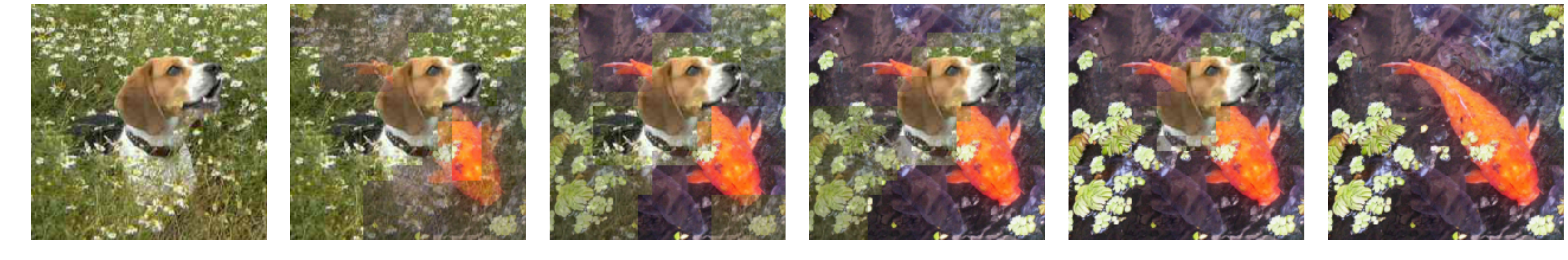}
\caption{Puzzle Mix images with increasing mixing weight $\lambda$.}
\label{fig:weight}
\end{figure*}

\begin{figure*}[t]
\centering
\includegraphics[width=\textwidth]{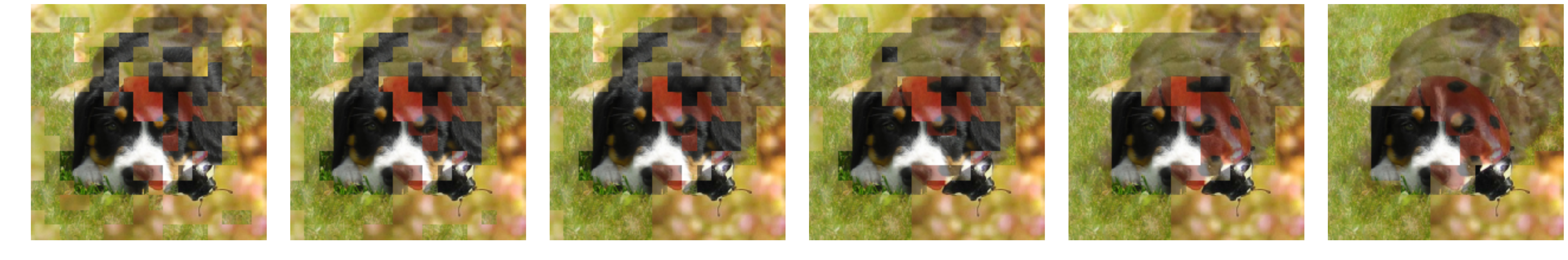}
\includegraphics[width=\textwidth]{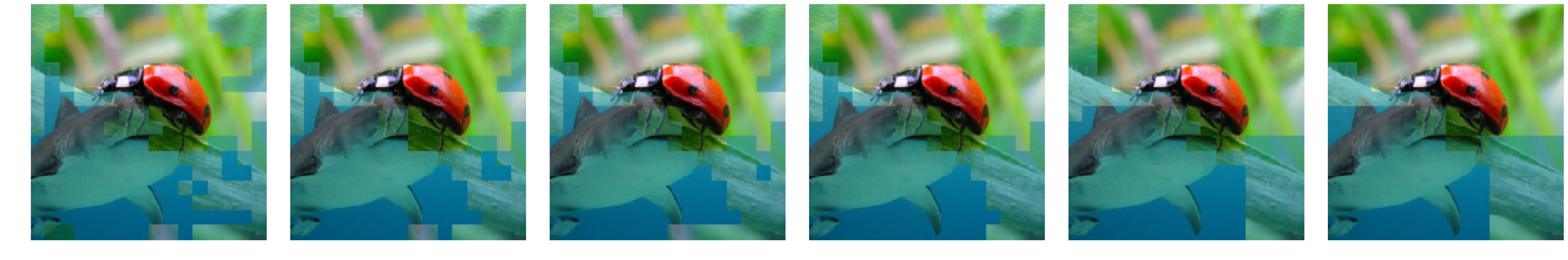}
\includegraphics[width=\textwidth]{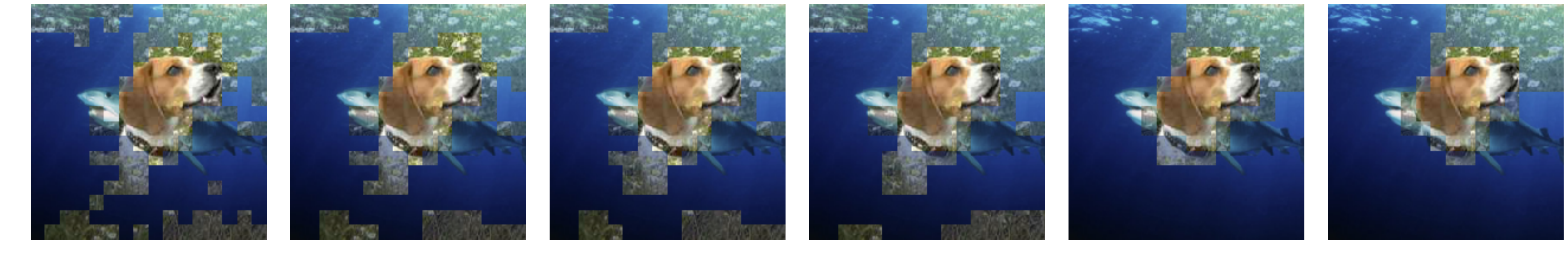}
\includegraphics[width=\textwidth]{figures/smooth/smooth4.pdf}
\includegraphics[width=\textwidth]{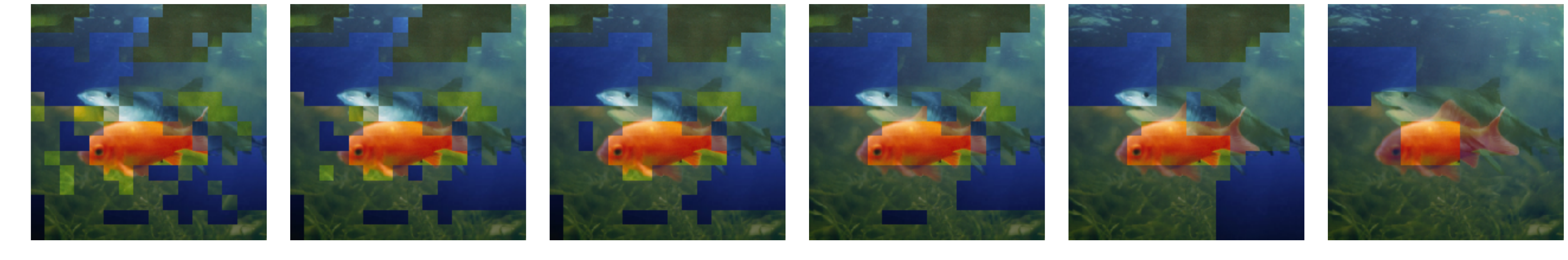}
\includegraphics[width=\textwidth]{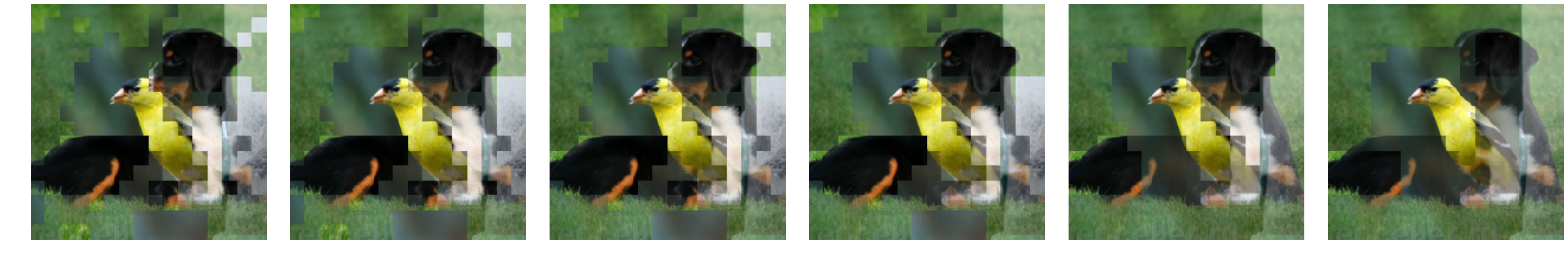}
\includegraphics[width=\textwidth]{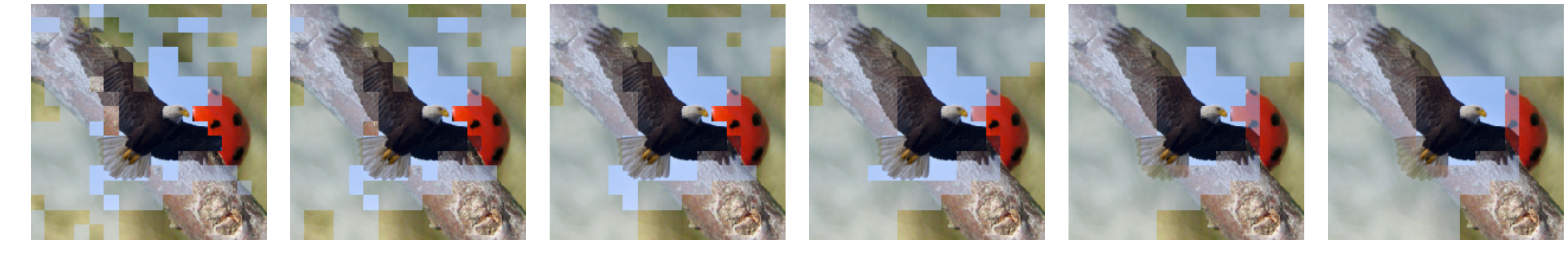}
\includegraphics[width=\textwidth]{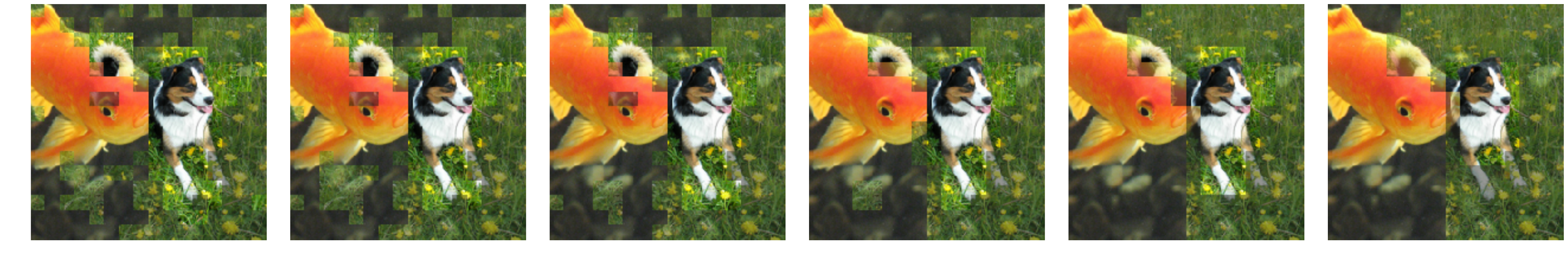}
\caption{Puzzle Mix images with increasing smoothness coefficient $\beta$ and $\gamma$.}
\label{fig:smooth}
\end{figure*}

In this section, we provide Puzzle Mix results while adjusting the hyperparameters associated with the optimal mask. 
In \Cref{fig:weight}, we visualize how the Puzzle Mix images change by increasing the mixing weight $\lambda$, and in \Cref{fig:smooth}, we visualize how the results change as we increase the coefficients of the smoothness terms, $\beta$ and $\gamma$.

\subsection{More Samples}
\begin{figure*}[t]
\centering
\includegraphics[width=0.49\textwidth]{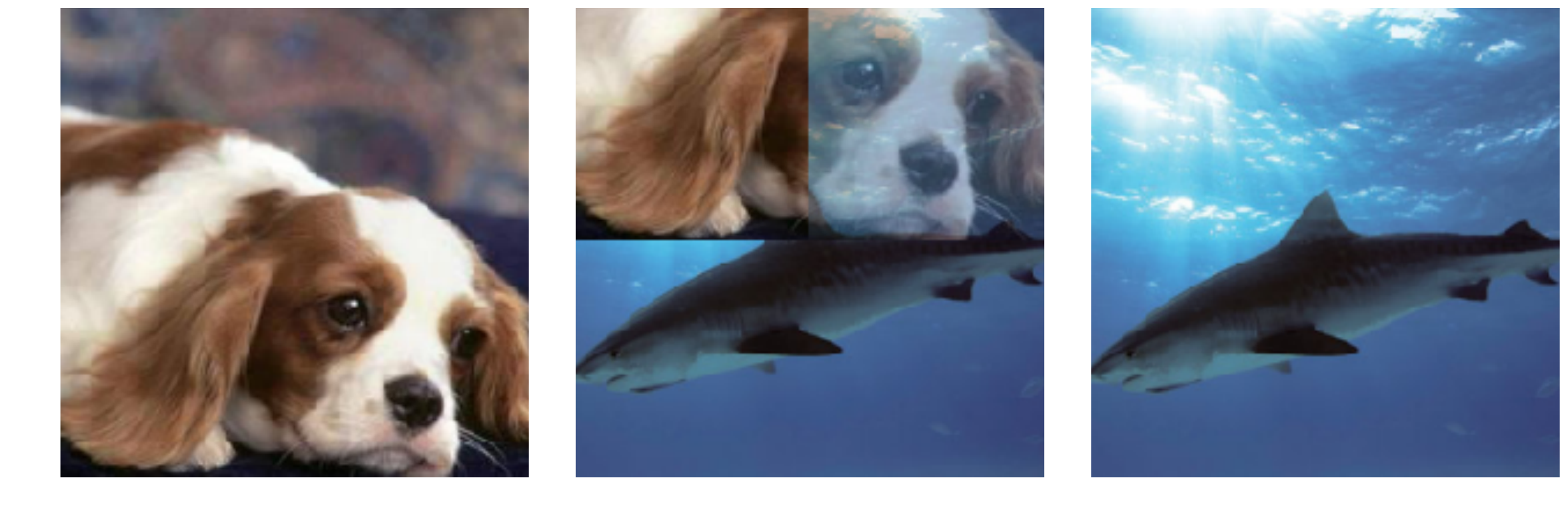}
\hspace{0.1cm}
\includegraphics[width=0.49\textwidth]{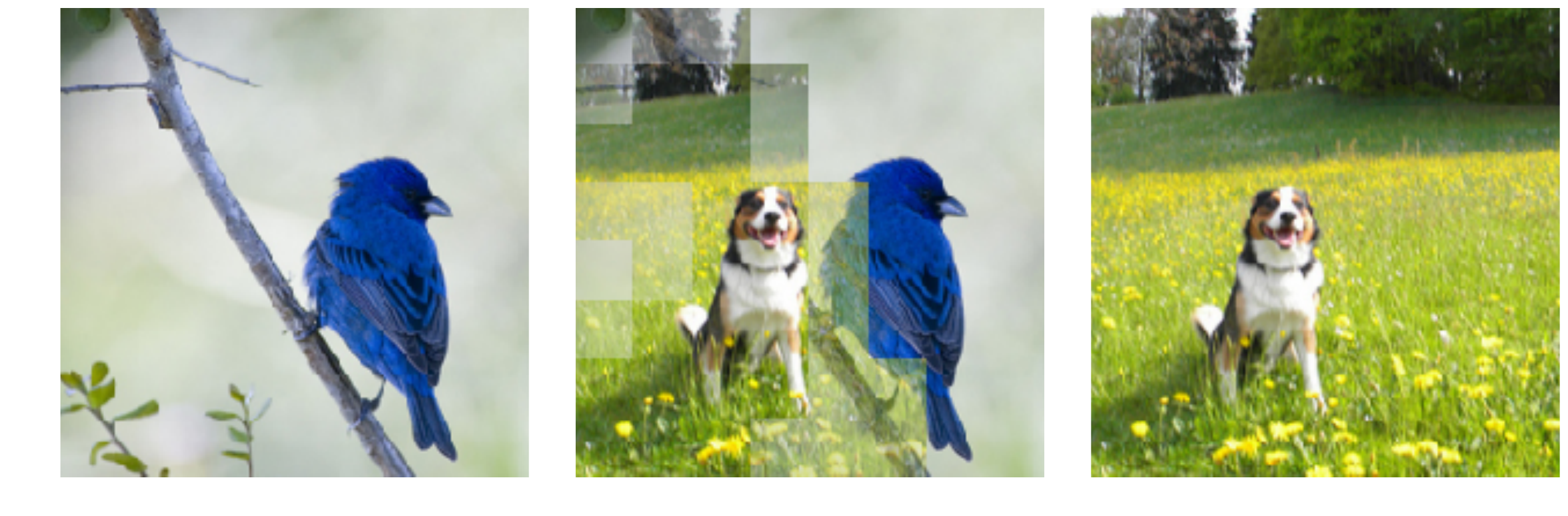}

\includegraphics[width=0.49\textwidth]{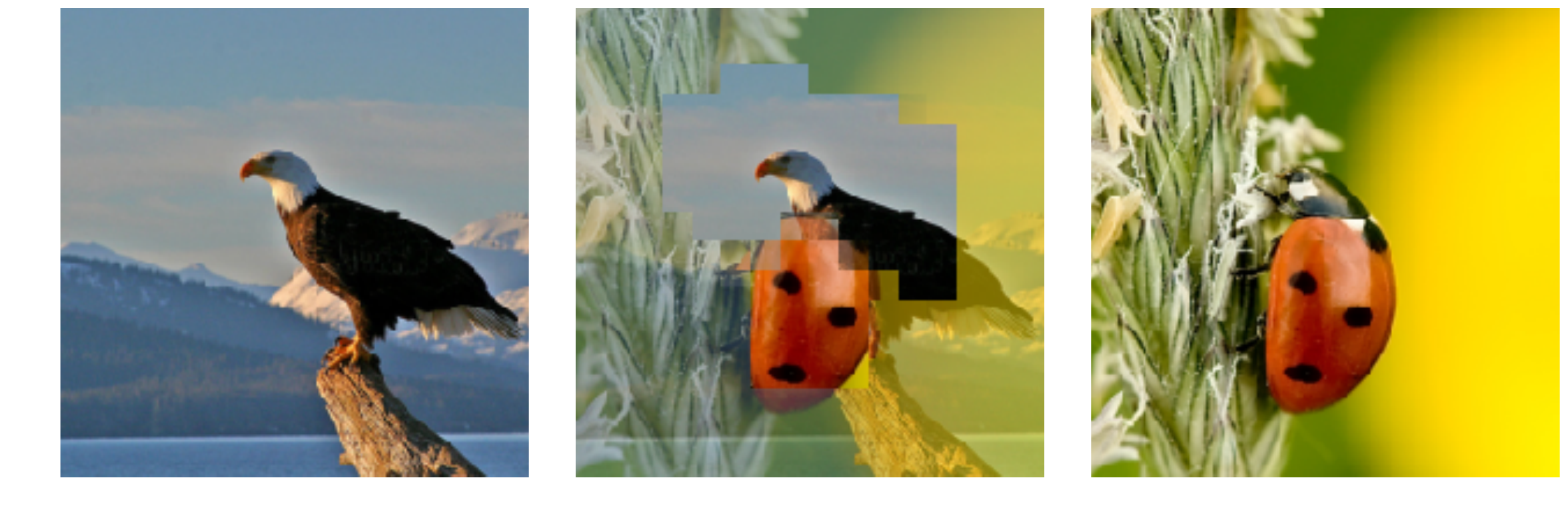}
\hspace{0.1cm}
\includegraphics[width=0.49\textwidth]{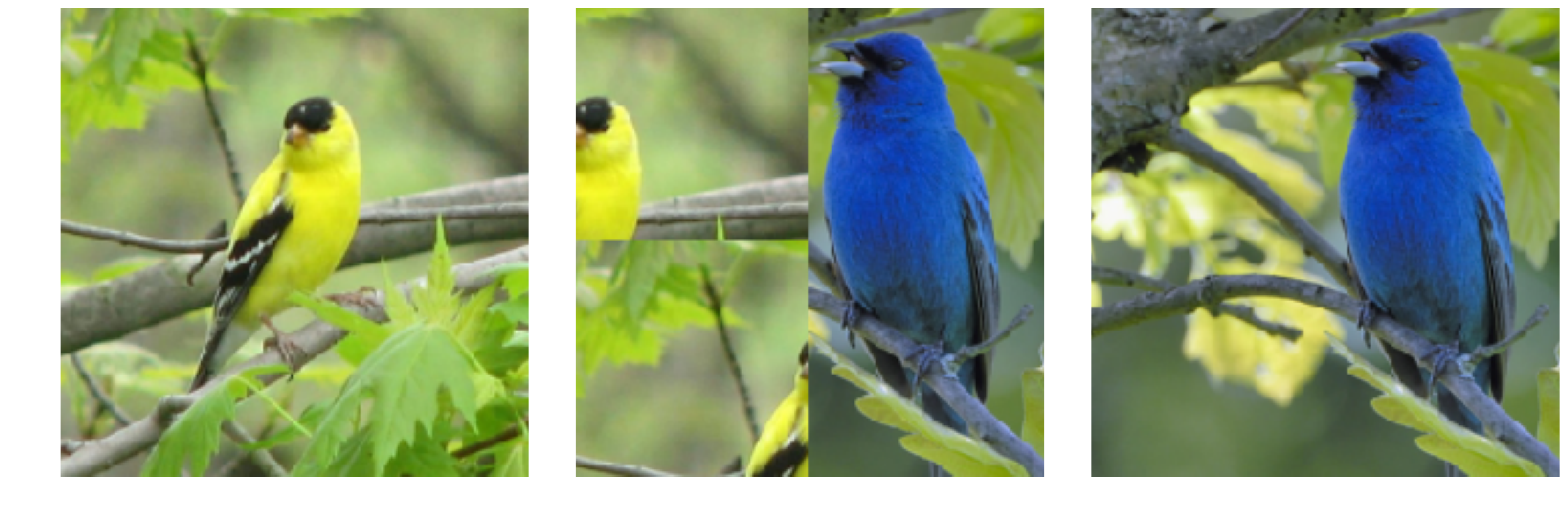}

\includegraphics[width=0.49\textwidth]{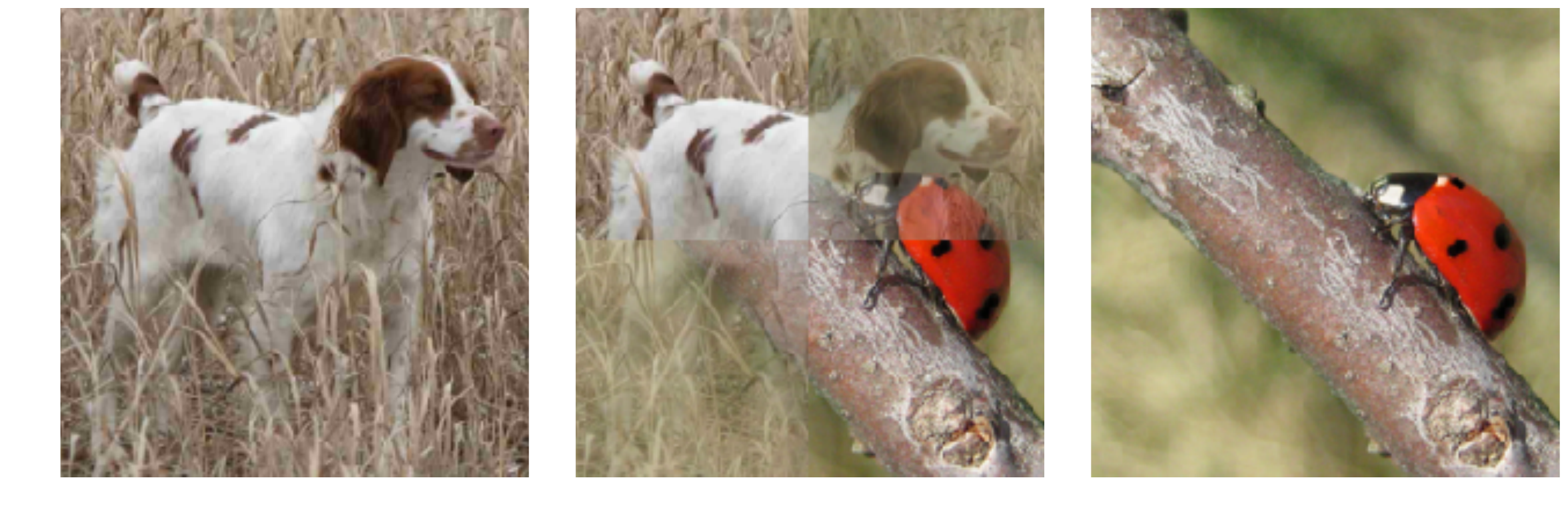}
\hspace{0.1cm}
\includegraphics[width=0.49\textwidth]{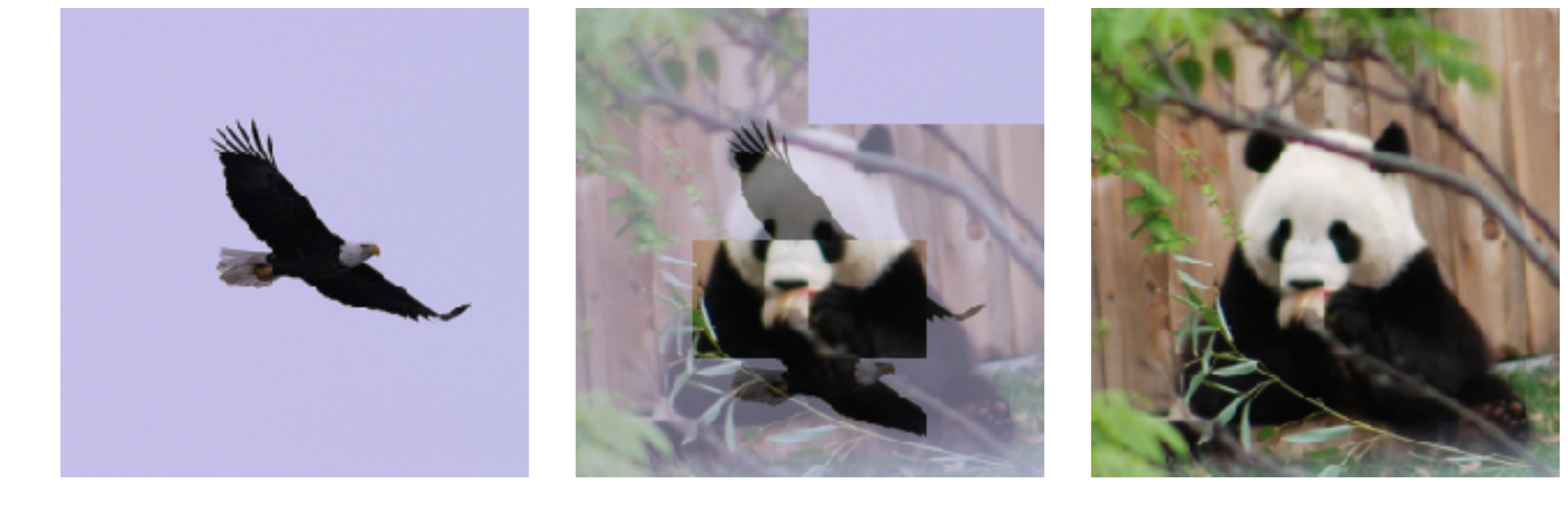}

\includegraphics[width=0.49\textwidth]{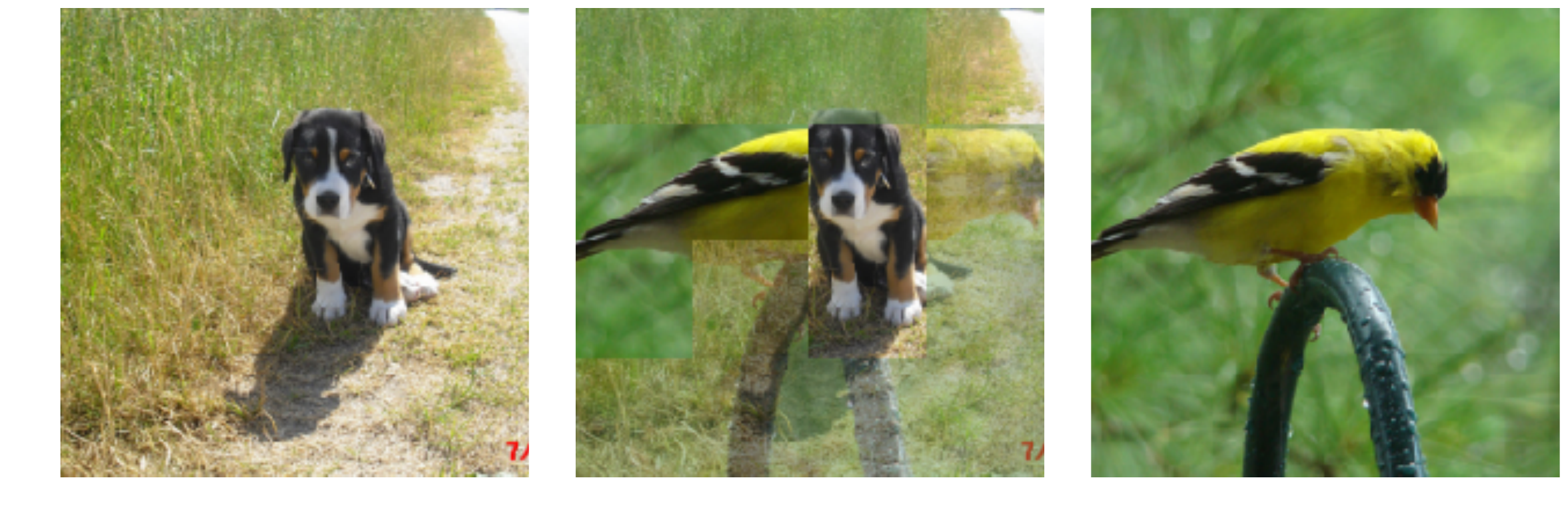}
\hspace{0.1cm}
\includegraphics[width=0.49\textwidth]{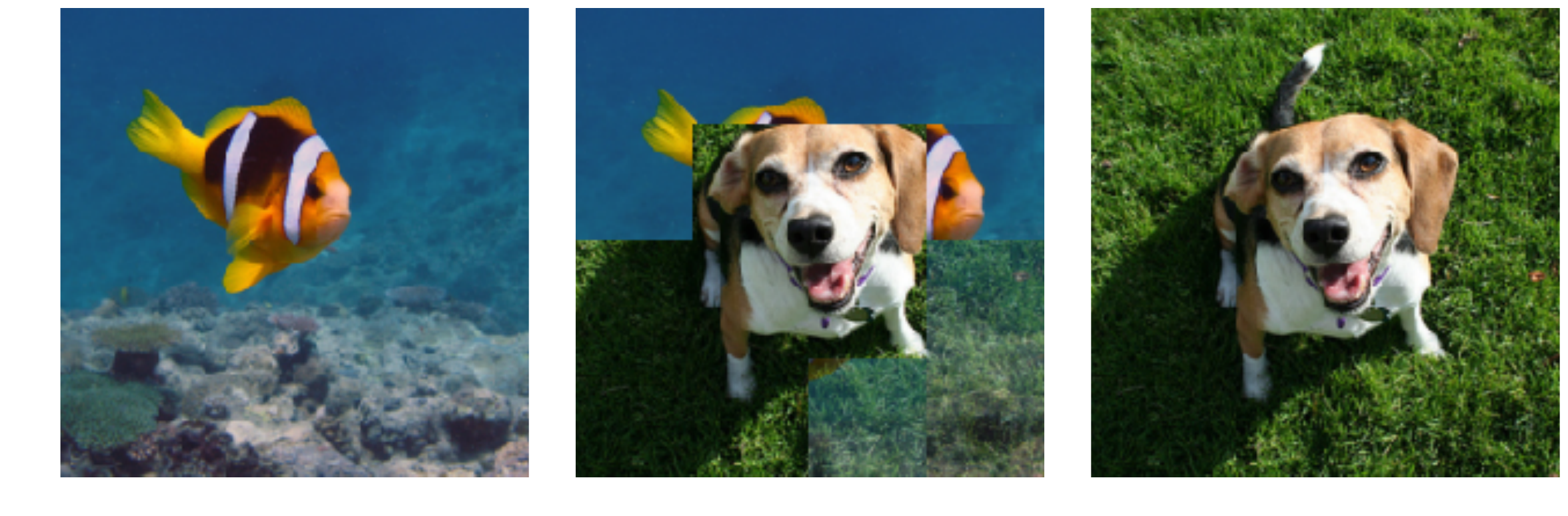}

\includegraphics[width=0.49\textwidth]{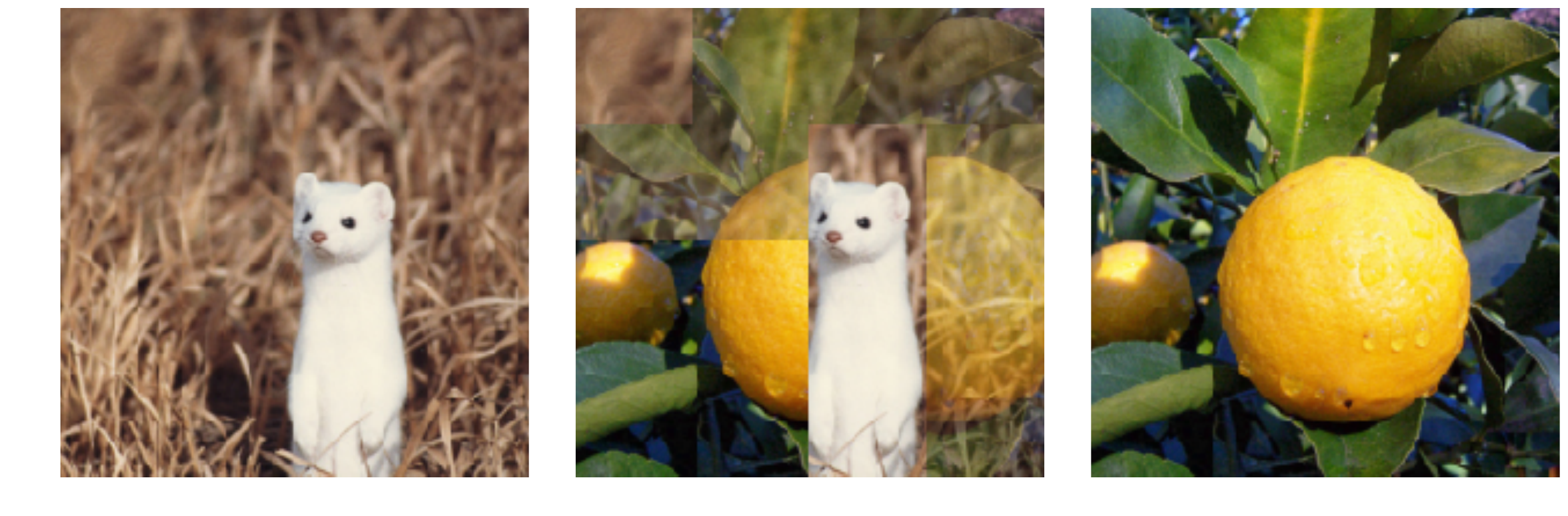}
\hspace{0.1cm}
\includegraphics[width=0.49\textwidth]{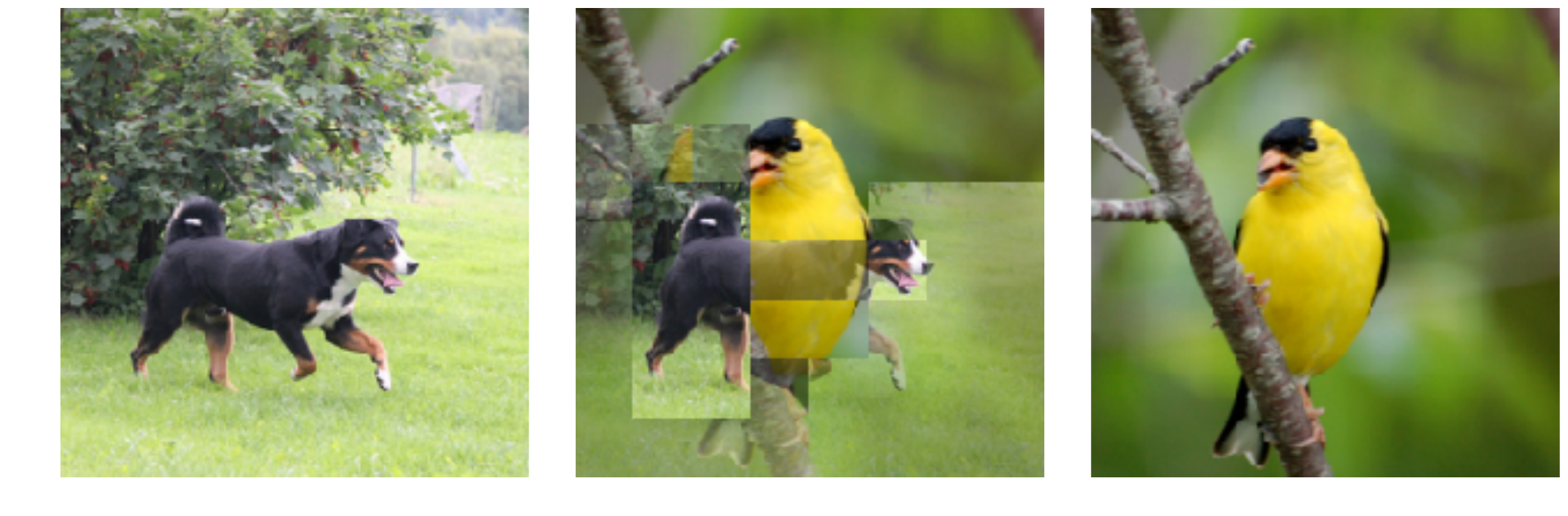}

\includegraphics[width=0.49\textwidth]{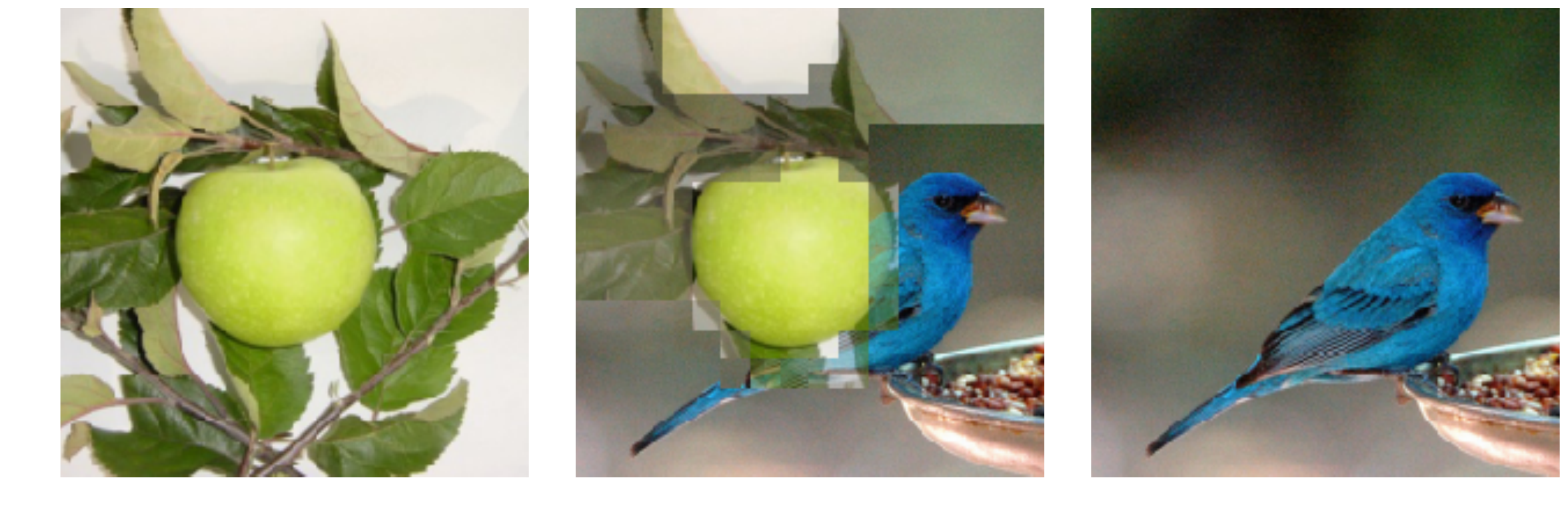}
\hspace{0.1cm}
\includegraphics[width=0.49\textwidth]{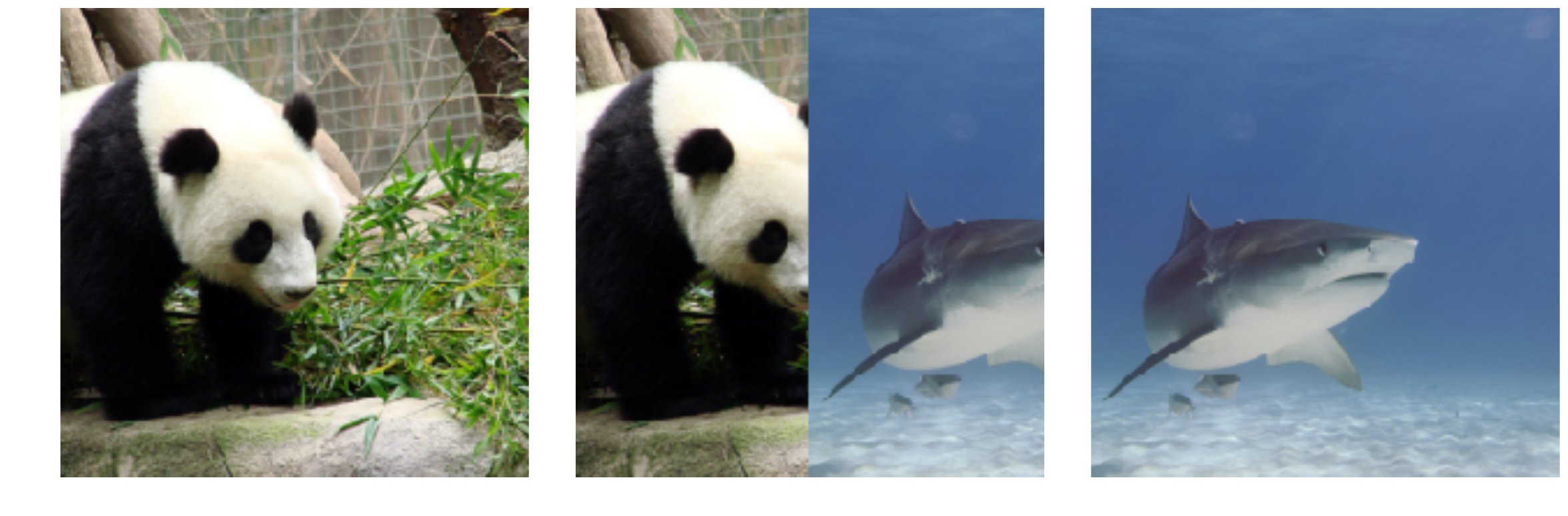}

\includegraphics[width=0.49\textwidth]{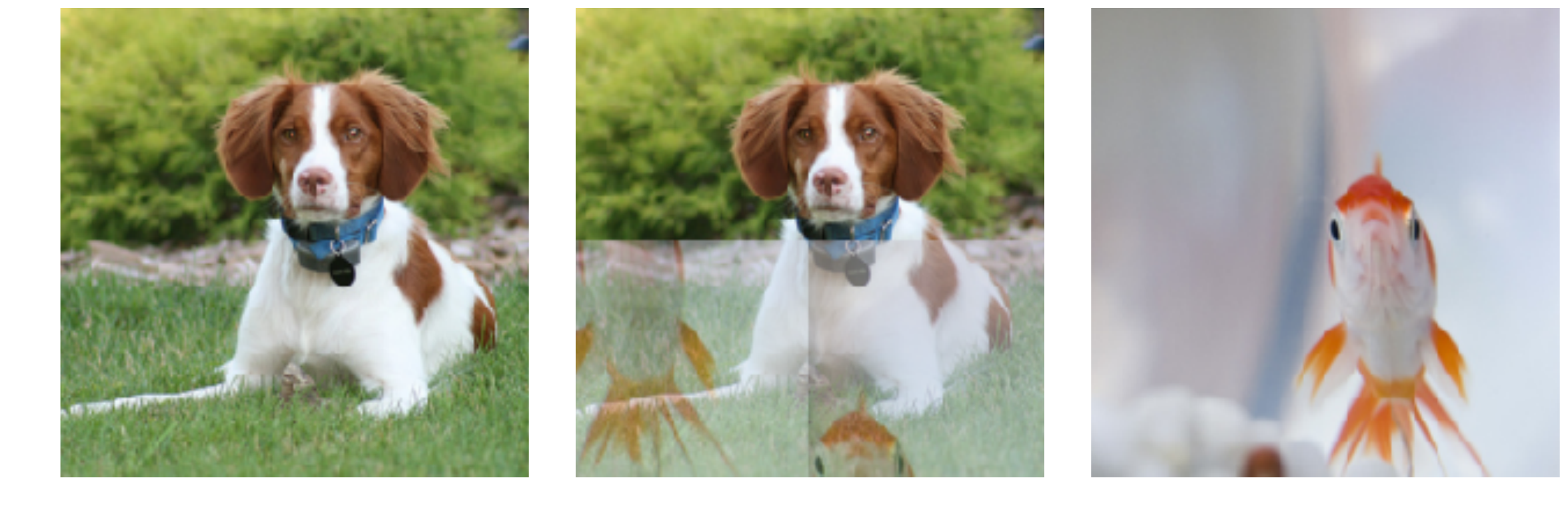}
\hspace{0.1cm}
\includegraphics[width=0.49\textwidth]{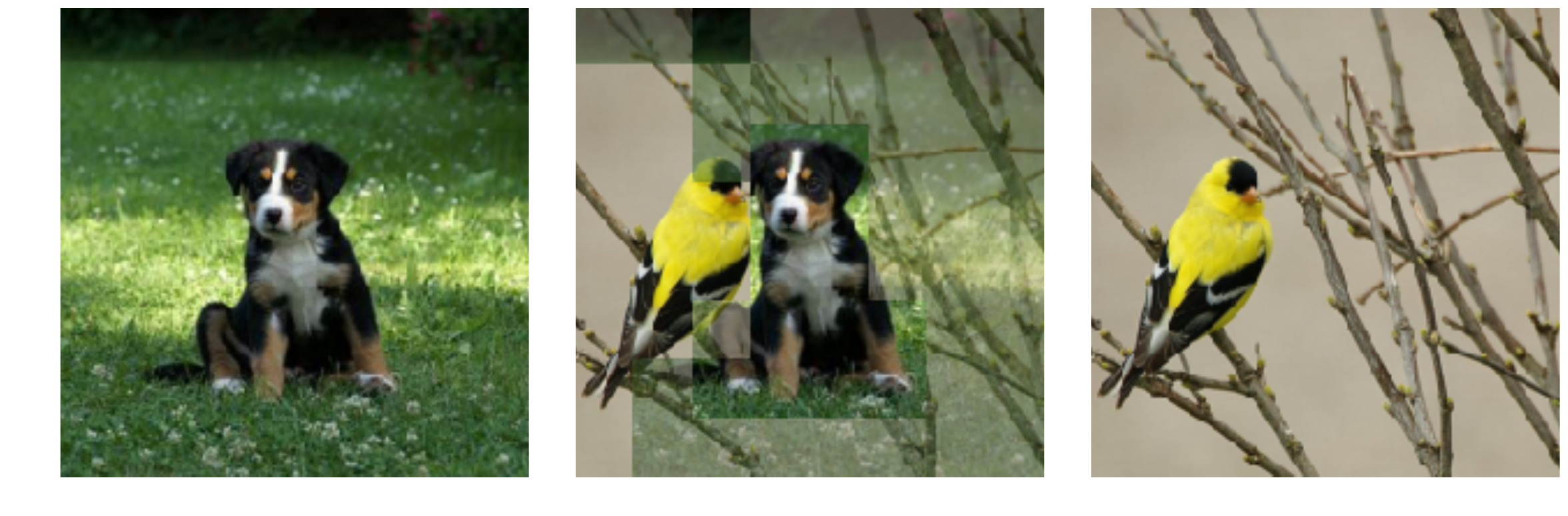}

\caption{Various Puzzle Mix image samples. Each row consists of input image 1 (left), Puzzle Mix image (middle), and input image 2 (right).}
\label{fig:final}
\end{figure*}

In this section, we provide Puzzle Mix results with various resolutions of the optimal mask and transport. \Cref{fig:final} visualizes the Puzzle Mix results along with the given inputs. 

\bibliography{main}
\bibliographystyle{styles/icml2020}

\end{document}